\documentclass[journal]{IEEEtran}

\IEEEoverridecommandlockouts

\usepackage[cmex10]{amsmath}
\usepackage{amssymb}
\usepackage{mdframed}
\usepackage{amsthm}
\usepackage{cases}
\usepackage{multirow}
\usepackage{empheq}

\usepackage{siunitx}

\usepackage[noadjust]{cite}
\usepackage{verbatim}

\usepackage{algorithm}
\usepackage{algpseudocode}

\ifCLASSINFOpdf
\usepackage[pdftex]{graphicx}
\else
\usepackage[dvips]{graphicx}
\fi
\usepackage[caption=false,font=footnotesize]{subfig}

\usepackage{color}
\usepackage{multirow}
\usepackage[hyphens]{url}

\begin{document}
	
    \sloppy
	
    \title{Collaborative Edge-to-Server Inference for Vision-Language Models}
	
    \author{
		\IEEEauthorblockN{Soochang~Song and Yongjune~Kim,~\IEEEmembership{Member,~IEEE}}	\\
    
\thanks{
    A preliminary version of this work is currently under review at the IEEE Global Commun. Conf. (GLOBECOM), 2026~\cite{Song2026communication}.
    S. Song and Y. Kim are with the Department of Electrical Engineering, Pohang University of Science and Technology (POSTECH), Pohang 37673, South Korea (e-mail: \{ssc6351, yongjune\}@postech.ac.kr).     
    }
    }
	
    \maketitle
    	
    \begin{abstract}
    We propose a collaborative edge-to-server inference framework for vision-language models (VLMs) that reduces communication cost while maintaining inference accuracy. 
    In typical deployments, visual data captured at edge devices (clients) is transmitted to the server for VLM inference.
    However, transmitting full-resolution images incurs high communication cost.
    Conversely, aggressive downsizing or excessive compression to mitigate communication overhead can discard fine-grained details, leading to accuracy degradation. 
    To overcome this limitation, we design a communication-efficient two-stage framework.
    In the first stage, the server performs inference on the downsized thumbnail (global image) and quantifies the min-entropy of the output tokens. 
    If the min-entropy exceeds a predefined threshold, the server identifies a region of interest (RoI) using the VLM's internal attention and requests the edge device to send a detail-preserved local image of the RoI.
    The server then refines its inference by jointly leveraging the global and local images.
    This selective retransmission strategy ensures that only essential visual content is additionally transmitted.
    Experimental results consistently confirm that the proposed framework substantially reduces communication overhead while maintaining inference accuracy across diverse VQA benchmarks.   
    \end{abstract}  

    \begin{IEEEkeywords}
        Edge-to-server inference, collaborative inference, multimodal semantic communications, entropy-aware image retransmission, vision-language models (VLMs). 
    \end{IEEEkeywords}
    
    \section{Introduction}
    
    With the rapid advancement of artificial intelligence (AI), the integration of multiple data modalities--such as images, text, and audio--into a shared embedding space has become a central paradigm in modern AI systems~\cite{Radford2021learning, Zhang2024vision, Liu2024improved}.
    Among these, vision-language models (VLMs), also referred to as multimodal large language models (MLLMs), have emerged as a prominent and widely adopted architecture, combining a vision encoder with a large language model (LLM) to enable visual reasoning capabilities~\cite{Liu2024improved, Dai2023InstructBLIP, Bai2025qwen2}.
    This architecture facilitates complex multimodal tasks that require joint understanding of visual and textual inputs, including visual question answering (VQA), image captioning, and image-text retrieval~\cite{Sharshar2025vision}.
    
    \begin{figure}[t!] 
        \centering
        \includegraphics[width=0.4\textwidth]{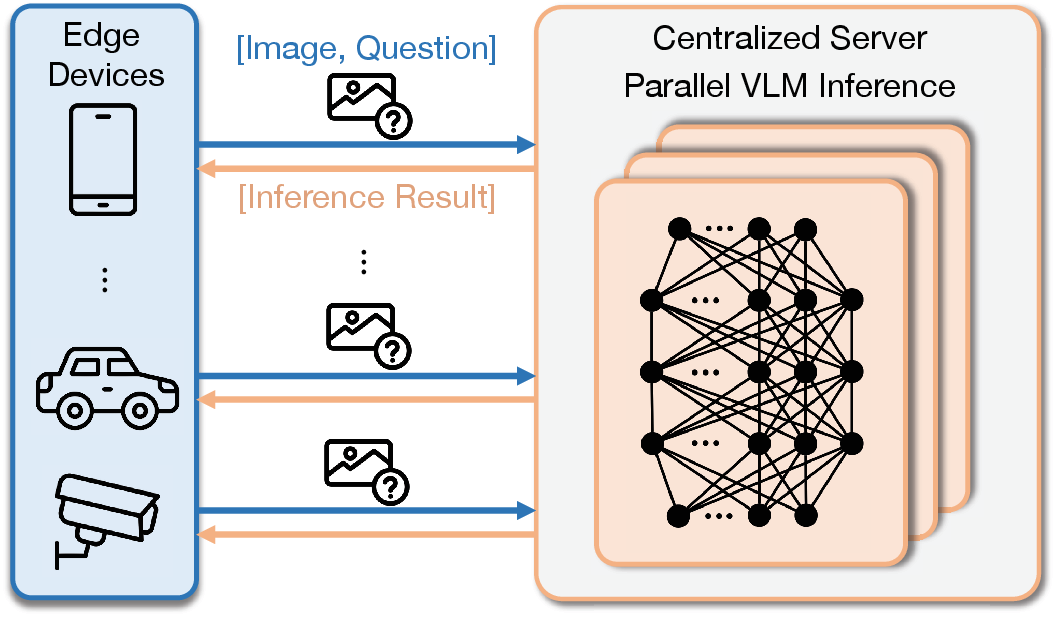}
        \caption{Schematic overview of the edge-to-server  VLM inference scenario.} 
        \label{fig:edge_server}
        \vspace{-4mm}
    \end{figure}    
    
    In real-world applications, as illustrated in Fig.~\ref{fig:edge_server}, heterogeneous edge devices (clients) often require VLM inference on locally captured visual data.
    However, deploying a full VLM architecture or its individual modules (e.g., a vision encoder) across diverse edge devices is generally impractical due to hardware heterogeneity.
    Consequently, visual data in the pixel domain, rather than processed visual tokens, is typically transmitted to a server hosting the VLM for inference.
    In this scenario, transmitting full-resolution images incurs high communication cost.
    Conversely, aggressive downsizing or excessive compression to mitigate communication overhead often discards essential fine-grained details, leading to accuracy degradation~\cite{Zhang2025mllms}.

    To address this problem, our framework introduces a two-stage entropy-aware retransmission (EAR) mechanism tailored to the collaborative edge-to-server setting. 
    In the first stage, the edge device transmits a low-resolution global image matching the input requirements of the VLM’s vision encoder (e.g., $336 \times 336$ pixels~\cite{Liu2024improved}), along with the question to the server. 
    The server performs an initial inference and quantifies its uncertainty using the min-entropy of the output tokens. 
    If the min-entropy is below a predefined threshold, the inference result is directly finalized. 
    Otherwise, a second stage is triggered: the server leverages the VLM’s internal attention scores to identify a task-relevant region of interest (RoI) and requests a detail-preserved local image from the edge device. 
    Upon receiving the retransmission request, the edge device extracts the local image from the original image and transmits it to the server. 
    The server then performs a second inference by jointly utilizing the visual tokens derived from both the global and local images, and evaluates the uncertainty of the second inference. 
    By comparing the uncertainties of the initial and refined inferences, the server adaptively selects the final answer from the two inference results.
    This uncertainty-aware, two-stage strategy reduces communication overhead by transmitting task-relevant high-quality visual details only when necessary, while maintaining inference accuracy across diverse tasks.
       
    At the core of the proposed framework is an entropy-aware decision mechanism that determines whether retransmission is required based on the VLM's inference uncertainty. 
    Since the server cannot directly ascertain whether the initial inference is correct, uncertainty is estimated from output probabilities and used as a proxy. 
    Specifically, we compute the per-token min-entropy from the softmax outputs of the LLM decoder at each generation step and average it across all output tokens to obtain a single scalar decision statistic. 
    This enables the server to request a local image only when the initial inference is likely unreliable, thereby attaining accuracy comparable to unconditionally transmitting local details while substantially reducing communication overhead.

    The proposed framework provides distinctive advantages depending on the vision encoder characteristics of the target VLMs.
    For the resolution-constrained model~\cite{Liu2024improved}, it enhances the model's effective capacity by mitigating the intrinsic information loss caused by input-layer downsampling~\cite{Radford2021learning,Zhang2025mllms}, thereby recovering fine-grained semantic details.
    On the other hand, for the recent models supporting variable input resolutions~\cite{Bai2025qwen2}, the proposed framework allocates communication resources more effectively by focusing on task-relevant image regions while suppressing redundant background information, thereby substantially reducing communication cost.
    We additionally note that our approach can be seamlessly integrated with traditional image codecs (e.g., JPEG) to further enhance communication efficiency.

   The main contributions of our work are summarized as follows:
    \begin{itemize}
    \item We propose a novel collaborative edge-to-server inference framework that achieves high inference accuracy and communication efficiency by jointly utilizing both global and local images. This framework adaptively balances the tradeoff between perception performance and transmission overhead by complementing a coarse global view with a task-relevant local view only when necessitated by the model's uncertainty.
    \item We introduce three core technical components: \emph{entropy-aware retransmission (EAR)} and \emph{entropy-guided selective refinement (SR)} for entropy-aware inference control, and \emph{attention-guided collaborative visual cropping} for spatial localization. These mechanisms leverage the model's internal attention to identify salient RoIs and utilize the min-entropy of prediction outputs to manage inference reliability, thereby ensuring robust and efficient VLM execution.
    \item The proposed method provides distinctive advantages depending on the VLM's architecture and does not require retraining or alignment between the VLM components, making it training-free and readily applicable to Internet of Things (IoT) environments with heterogeneous edge-computing resources.
    \end{itemize}

    Experiments on various VQA tasks show that the proposed framework preserves the accuracy of ViCrop~\cite{Zhang2025mllms} while reducing additional communication costs ranging from \SI{24}{\%} to \SI{76}{\%} depending on the task complexity and model configuration.
    These results confirm that the integration of entropy-aware retransmission and attention-guided collaborative visual cropping enables a highly adaptive inference process.
    Our approach effectively optimizes the tradeoff between communication resources and task performance, validating its practical applicability for edge-to-server VLM deployments.

    The rest of this paper is organized as follows. 
    Section~\ref{sec:background} provides an overview of the VLM architecture with the attention mechanism and the attention-guided visual cropping scheme, followed by a review of related literature in Section~\ref{sec:related}
    Section~\ref{sec:system} details our edge-to-server inference framework. 
    The core components--entropy-aware retransmission, attention-guided collaborative visual cropping, and entropy-guided selective refinement--are detailed in Sections~\ref{sec:entropy}, \ref{sec:collaborative_vicrop}, and \ref{sec:selective_refinement}, respectively.
    Section~\ref{sec:results} provides experimental results, followed by conclusions in Section~\ref{sec:conclusion}.

    \section{Background}\label{sec:background}

    \subsection{Vision-Language Models}\label{sec:vlm}    

    \begin{figure}[t!] 
        \centering
        \includegraphics[width=0.45\textwidth]{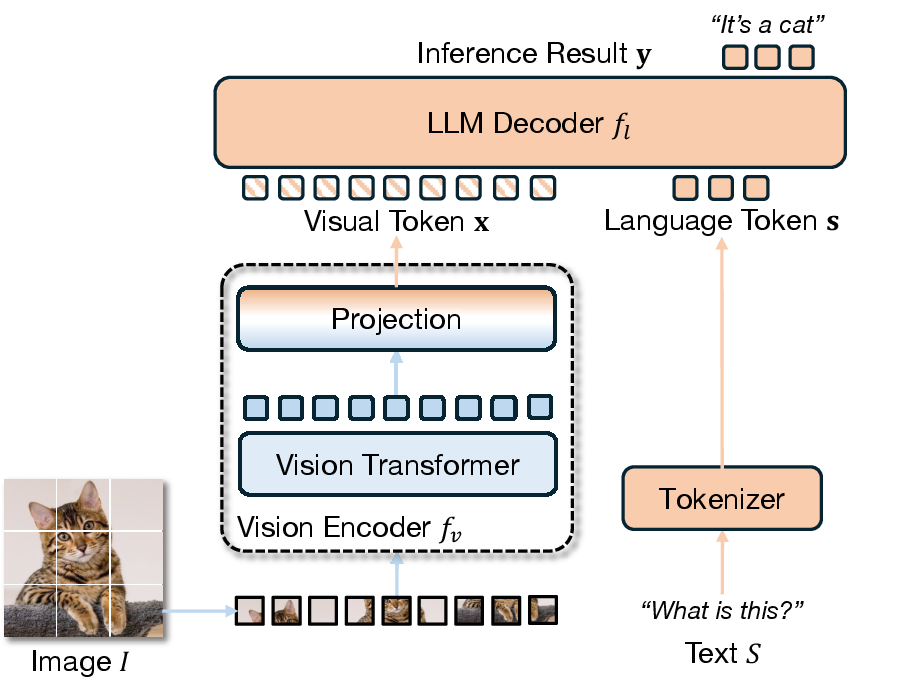}
        \caption{Overview of the VLM architecture~\cite{Liu2024improved}.} 
        \label{fig:vlm}
        \vspace{-4mm}
    \end{figure}

    VLMs integrate visual and linguistic inputs within a unified transformer architecture and typically consist of two components: a vision encoder~\cite{Radford2021learning} and an LLM decoder~\cite{Touvron2023llama, Mesnard2024gemma, Chiang2023vicuna}.
    A simplified overview of the VLM architecture is shown in Fig.~\ref{fig:vlm}. 

    The model takes as input an image–text pair $(I, S)$, where $I \in \mathbb{R}^{h \times w \times 3}$ denotes an original RGB image in the pixel domain and $S$ denotes a natural language sentence (e.g., a question referring to the image in a VQA task). 
    Here, $h$ and $w$ represent the height and width of the original image, respectively.
    The image $I$ is first resized to $\hat{I} \in \mathbb{R}^{\hat{h} \times \hat{w} \times 3}$ to match the input resolution of the vision encoder, where $\hat{h}$ and $\hat{w}$ denote the height and width of the encoder's input. 
    The vision encoder $f_v(\cdot)$ consists of two modules: a Vision Transformer (ViT)~\cite{Dosovitskiy2021image, Radford2021learning} and a multimodal projection layer~\cite{Liu2024improved, Dai2023InstructBLIP}.
    The ViT divides the image into square patches and extracts visual features. 
    The multimodal projector then maps these features into the shared embedding space of the LLM by producing visual tokens $\mathbf{x} \in \mathbb{R}^{n_v \times d_{\text{LLM}}}$, where $n_v$ and $d_{\text{LLM}}$ denote the number of visual tokens and the hidden dimension of the LLM, respectively. 
    The visual encoding process can be expressed as 
    \begin{equation}
        \mathbf{x} = f_v(\hat{I}).
    \end{equation}
    
    The text input $S$ is tokenized using a pretrained tokenizer~\cite{Kudo2018sentencepiece}, yielding a sequence of language tokens:
    \begin{equation}
        \mathbf{s} = \operatorname{Tokenize}(S),
    \end{equation}
    where $\mathbf{s} \in \mathbb{R}^{n_s \times d_{\text{LLM}}}$ and $n_s$ denotes the number of language tokens.
    Given the visual tokens $\mathbf{x}$, language tokens $\mathbf{s}$, and previously generated outputs $\mathbf{y}_{<t}$, the LLM decoder $f_l(\cdot)$ generates the output token $y_t$ autoregressively:
    \begin{equation}
        y_t = f_l(\mathbf{x}, \mathbf{s}, \mathbf{y}_{<t}),
    \end{equation}
    where $y_t \in \mathbb{R}^{d_{\text{LLM}}}$ represents the output token generated at step $t$.
    The final sequence $\mathbf{y} = \{y_1, \dots, y_T\}$ is then detokenized~\cite{Kudo2018sentencepiece} to produce the natural language output, where $T$ denotes the total number of decoding steps. 
    
    \subsection{Attention Mechanism}\label{sec:attention}   
    
    Transformer layers in both the visual encoder and LLM decoder rely on the multi-head self-attention (MHSA) mechanism, though there are subtle differences in how it is applied.
    Given an input sequence \( \mathbf{z} \in \mathbb{R}^{n \times d} \), where $n$ denotes the sequence length and $d$ denotes the transformer dimension. 
    The query, key, and value matrices are computed through linear projections:
    \begin{equation}
    [Q, K, V] = \mathbf{z} U_{QKV}, \quad U_{QKV} \in \mathbb{R}^{d \times 3d_H},
    \end{equation}
    where $Q, K, V \in \mathbb{R}^{n \times d_H}$, and $d_H = d / H$ denotes the per-head dimension with $H$ attention heads.
    
    In the ViT, unmasked self-attention is applied, and the attention weight matrix is computed as:
    \begin{equation}
        A = \text{softmax}\left( \frac{Q K^\top}{\sqrt{d_H}} \right).
    \end{equation}
    In contrast, the LLM decoder employs masked self-attention~\cite{Vaswani2017attention} to preserve the autoregressive property of language modeling. 
    An additive causal mask $ M \in \mathbb{R}^{n \times n} $ is applied to prevent tokens from attending to future positions:
    \begin{equation}
        A = \text{softmax}\left( \frac{Q K^\top}{\sqrt{d_H}} + M\right).
    \end{equation}
    The self-attention output is then given by
    \begin{equation}
        \text{SA}(\mathbf{z}) = AV.
    \end{equation}
    The MHSA mechanism extends this operation by computing $H$ parallel self-attention outputs and concatenating them:
    \begin{equation}
        \text{MHSA}(\mathbf{z}) = \left[\text{SA}_1(\mathbf{z}), \cdots, \text{SA}_H(\mathbf{z})\right] U_{\text{MHSA}},
    \end{equation}
    where $U_{\text{MHSA}} \in \mathbb{R}^{d \times d}$ is the projection matrix for the MHSA output.
    
    The attention weight matrix $A$ provides interpretability by indicating which input tokens the model focuses on when responding to a given query.
    The choice of query for attention analysis depends on the task and the model architecture. 
    For image classification using ViTs, the class token typically serves as the query~\cite{Dosovitskiy2021image}.
    In contrast, for language modeling in LLMs or VLMs, the generated output tokens can serve as the queries that attend to the encoded inputs~\cite{Chen2024image,Zhang2025beyond, Zhang2025mllms}.

    \subsection{Attention-Guided Visual Cropping}\label{sec:ViCrop}   
    Attention-guided visual cropping (ViCrop)~\cite{Zhang2025mllms} is an effective strategy for localizing task-relevant RoI by leveraging the VLM's internal attention mechanisms.
    A primary finding in \cite{Zhang2025mllms} is that state-of-the-art MLLMs often exhibit a perceptual bias toward coarse visual features, thereby failing to capture fine-grained details essential for complex reasoning.
    To address this limitation, ViCrop provides a training-free method for automatic visual localization based on the model’s attention dynamics.  
    A key intuition behind this approach is that VLMs tend to focus on semantically meaningful regions even when generating incorrect responses, providing an implicit representation of where the model \emph{looks}. 
    By re-encoding the identified RoI from the original high-resolution image as an auxiliary input, ViCrop enhances the model's perceptual precision.

    To isolate the most salient regions, a relative attention map, $A_{\text{rel}}$, is derived to suppress systematic attention artifacts~\cite{Darcet2024vision} and accentuate task-specific semantic relevance.
    Specifically, the attention weights from the LLM decoder’s starting output token directed toward image tokens are extracted, yielding an attention tensor $A_{\text{si}}(\mathbf{x}, \mathbf{s}) \in \mathbb{R}^{L \times H \times 1 \times n_v}$.  
    The attention values are averaged across all $H$ heads, and a specific layer $l \in \{1, \ldots, L\}$ is empirically selected to maximize the contrast between high- and low-attention regions, yielding $\hat{A}_{\text{si}}^l(\mathbf{x}, \mathbf{s}) \in \mathbb{R}^{1 \times n_v}$.  
    To eliminate prompt-agnostic artifacts, the attention for the task-specific query $\mathbf{s}$ is then normalized by that of a generic instruction $\mathbf{s}'$ (“Write a general description of the image.”):
    \begin{equation}
        A_{\text{rel}} = \frac{\hat{A}_{\text{si}}^l(\mathbf{x}, \mathbf{s})}{\hat{A}_{\text{si}}^l(\mathbf{x}, \mathbf{s'})} \in \mathbb{R}^{1 \times n_v}.
    \end{equation}
    Based on $A_{\text{rel}}$, the final RoI is determined via a sliding-window search across various candidate scales (e.g., $\{1.0, 1.2, \dots, 2.0\}$ times the input resolution of the vision encoder) over the relative attention map.
    For each candidate, the sum of relative attention values within the corresponding window is computed, and the region exhibiting the highest internal-external contrast is selected as the final bounding box for cropping.
    While adapting the ViCrop mechanism into our framework, we decouple functional execution between the edge and server, ensuring its activation is selectively triggered only for the samples with high uncertainty.
    
    \section{Related Work}\label{sec:related}

    \subsection{Multimodal Semantic Communications}
    
    With the rapid advancement of AI, the paradigms of semantic and task-oriented communications, which focus on extracting and transmitting only task-relevant information, have garnered significant attention.
    Unlike conventional approaches that focus on bit- or symbol-level fidelity, these approaches aim to maintain semantic fidelity and task performance~\cite{Gunduz2022beyond, Shi2021from, Lan2021what, Huang2022toward, Xie2020lite, Liu2025goal}.
    Recent progress in multimodal AI has further expanded the potential of semantic communication.
    For instance,~\cite{Zhang2024Unified} proposes a transformer-based framework that independently encodes text, image, and audio modalities before feeding them into a unified decoder, thereby leveraging cross-modal relationships to improve communication efficiency.
    
    Several studies have explored the integration of VLMs into communication systems. 
    For example,~\cite{Yuan2025task} proposes a collaborative inference framework that deploys the vision encoder on an edge device and the language model decoder on a server.
    To reduce communication cost of high-resolution images, they employ vector quantization and neural compression on the extracted visual features.
    The authors of~\cite{Zhang2025vavlm} leverage VLM-derived RoI information to extract a subset of image regions from video frames, effectively reducing the bandwidth required for transmission. 
    While these approaches demonstrate the feasibility of integrating VLMs into communication systems, they often require non-trivial computation on edge devices. 
    In contrast, our framework performs most computations on the server, imposing minimal computational and power burden on the edge device.

    \subsection{Uncertainty Estimation in AI Inference}
    As AI models are increasingly deployed in critical domains such as biomedicine, security, and autonomous systems, assessing and mitigating the risks of hallucinations and erroneous predictions has become imperative~\cite{Huang2025survey}.
    To address this challenge, a growing body of research focuses on quantifying model uncertainty as a proxy for inference reliability.
    In natural language processing (NLP), token-level uncertainty estimation has been extensively investigated.
    For example,~\cite{Fadeeva2024fact, Neha2024language} quantify the uncertainty of a generated sentence based on the probability distribution over output tokens.
    Building upon these efforts, we propose an edge-to-server collaborative inference framework for multimodal tasks, in which uncertainty-aware retransmission effectively reduces communication cost while maintaining inference accuracy. 
    
    In the context of semantic communications,~\cite{Im2024attention} applies uncertainty estimation to collaborative edge-to-server inference for image classification.
    Specifically, the entropy of the softmax output from the ViT classification head is used to decide whether the prediction of a lightweight edge model should be accepted or whether the image should be transmitted to the server for more reliable inference.
    Beyond image classification, our work builds upon the underlying principle of uncertainty-aware collaborative inference and reformulates it for VLMs, introducing an uncertainty-aware retransmission strategy that effectively reduces communication cost for various VQA tasks. 

    \section{Collaborative Edge-to-Server Inference Framework for Vision-Language Models}\label{sec:system} 

    \begin{figure*}[t] 
        \centering
        \includegraphics[width=0.90\textwidth]{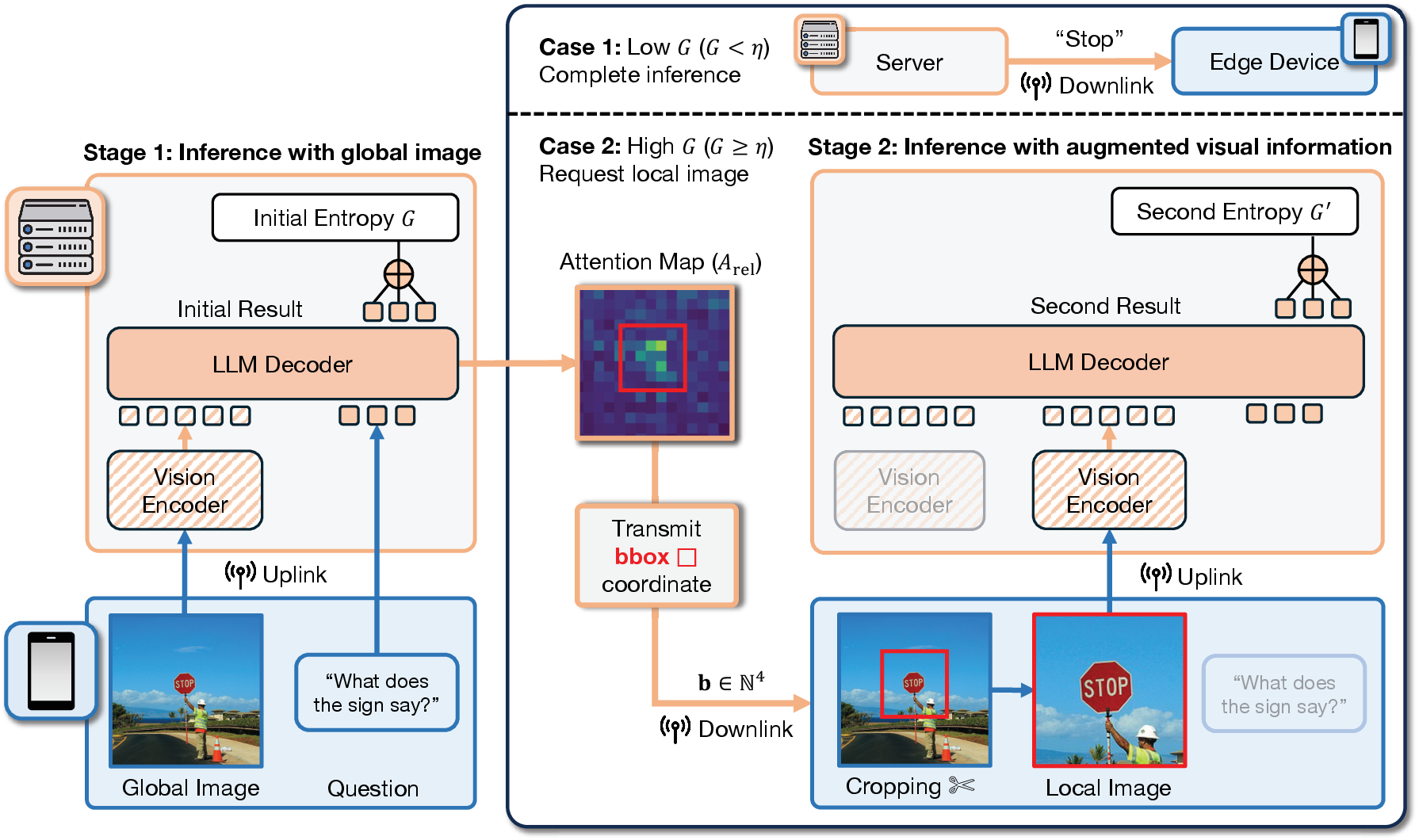}
        \caption{Overview of the proposed two-stage collaborative edge-to-server inference framework. The edge device first transmits a global image to the server for initial inference. If high uncertainty is detected in the output token distribution, the server requests a local image based on an attention-derived bounding box. The edge device then crops the corresponding region from the original image and transmits it to the server. The server performs a second inference using both global and local visual features.} 
        \label{fig:framework}
        \vspace{-4mm}
    \end{figure*}

    We propose a collaborative edge-to-server inference framework built on pre-trained, off-the-shelf VLMs.    
    The proposed framework incorporates entropy-aware image retransmission and attention-guided collaborative visual cropping inspired by~\cite{Zhang2025mllms} to reduce communication cost while maintaining inference accuracy.
    In this setting, each client uses an edge device connected to the server over a communication channel. 
    The inference task is performed on input data residing on the edge device, consisting of visual images and textual questions posed by the client. 
    Due to the hardware heterogeneity of edge devices, it is infeasible to deploy and execute the full VLM architecture or any of its components across diverse edge devices directly.
    Hence, the pixel-domain visual data should be transmitted to the server hosting the VLM.
    
    Under these circumstances, while transmitting images at full resolution ensures the preservation of critical details, it incurs a substantial increase in communication overhead.
    In contrast, employing heavy downsizing or high compression ratios to alleviate communication demand often results in the loss of fine-grained features, thereby compromising inference accuracy. 
    To address this problem, we introduce a two-stage image transmission and inference protocol between the edge device and the server.
    As shown in Fig.~\ref{fig:framework}, the proposed framework achieves communication efficiency by adaptively balancing communication cost and inference accuracy through uncertainty-aware retransmission and attention-guided refinement.

    In the proposed framework, the edge device first transmits a visual input (i.e., a global image obtained by downscaling the original image to match the vision encoder configuration) and a textual input (i.e., a question posed by the client) to the server.
    Upon receiving both inputs, the server-side VLM performs an initial inference.
    During inference, the server estimates the uncertainty of its prediction using the min-entropy of the output distribution of the server model. 
    If the min-entropy is low, indicating high confidence, the inference result is finalized and transmitted to the edge device.
    Conversely, if the min-entropy is high, a secondary stage is triggered.
    
    In the second stage, the server computes a relative attention map over the global image to identify the VLM's RoI.
    Based on this relative attention map, the server identifies the position of the bounding box that exhibits the highest attention activation.
    The bounding box coordinates in the original image are then transmitted to the edge device, along with a request for a detailed local image.
    The edge device crops the original image according to the received bounding box coordinate, generating a local image that contains high-quality details of semantically important regions.
    This local image is transmitted to the server, where it is concatenated with the tokens of the global image to perform a second inference with augmented visual information.
    During inference, the server evaluates inference uncertainty in the same manner as the initial stage.
    By comparing the min-entropy values of the first and second inferences, the server adaptively selects the most reliable result as the final output and transmits it to the edge device.
    The refined answer is finally returned to the edge device.
    
    \begin{algorithm}[t] 
        \caption{Proposed Collaborative Edge-to-Server Inference Framework} \label{algo:proposed}
        \textbf{Input:} An image and text pair $(I, S)$ on edge device. \\
        \textbf{Output:} Generated output $\mathbf{y}$.
        \begin{algorithmic}[1]
            \State $\hat{I} \gets \operatorname{Resize}(I)$ \Comment{Performed on edge}
            \State Transmit $\hat{I}$ and $S$ to the server
    
            \State $\mathbf{x} \gets f_v(\hat{I})$, \quad $\mathbf{s} \gets \operatorname{Tokenize}(S)$ \Comment{Initial encoding}
    
            \For {$t = 1:T$} \Comment{Initial inference stage}
                \State $y_t \gets f_l(\mathbf{x}, \mathbf{s}, \mathbf{y}_{<t}), \quad g_t \gets g(\mathbf{x}, \mathbf{s}, \mathbf{y}_{<t})$ 
            \EndFor
            \State $\mathbf{y} \gets \{y_1,\ldots y_T\}$ 
            \State $G \gets \frac{1}{T} \sum_{t=1}^T g_t$ \Comment{Obtain average initial entropy}
    
            \If {$G\ge \eta$}
                \State $A_{\text{rel}} \gets \operatorname{ComputeRelativeAttention}(\mathbf{x}, \mathbf{s})$ 
                \State $\mathbf{b} \gets\operatorname{FindBoundingBox}(A_{\text{rel}})$ 
                \State Request $\tilde{I}$ from edge device using $\mathbf{b}$
                \State $\tilde{I} \gets \operatorname{Resize}(\operatorname{Crop}(I, \mathbf{b}))$
                \Comment{Performed on edge}
                \State Transmit $\tilde{I}$ to the server
    
                \State $\tilde{\mathbf{x}} \gets f_v(\tilde{I})$ \Comment{Encode local image}
                \For {$t = 1:T'$}  \Comment{Second inference stage}
                    \State $y'_t \gets f_l(\mathbf{x}, \tilde{\mathbf{x}}, \mathbf{s}, \mathbf{y}_{<t}'), \quad g'_t \gets g(\mathbf{x}, \tilde{\mathbf{x}}, \mathbf{s}, \mathbf{y}'_{<t})$ 
                \EndFor
                \State $\mathbf{y'}{} \gets \{y'_1,\ldots y'_{T'}\}$ 
                \State $G' \gets \frac{1}{T'} \sum_{t=1}^{T'} g'_t$ \Comment{Obtain average second entropy}
                \If {$G' \le G + \delta$}   \Comment{Selective refinement}
                    \State $\mathbf{y} \gets \mathbf{y}'$ 
                \EndIf
            \EndIf
    
            \State \textbf{return} $\mathbf{y}$
        \end{algorithmic}
    \end{algorithm}

    The overall process is summarized in Algorithm~\ref{algo:proposed}.
    On the edge device, given an image–text pair $(I, S)$, the image is resized to match the VLM input resolution: $\hat{I} = \operatorname{Resize}(I)$. 
    The resized image $\hat{I}$ and the corresponding sentence $S$ are then transmitted to the server.
    Upon receiving these inputs, the server extracts the global visual feature $ \mathbf{x} = f_v(\hat{I})$ and encodes the input sentence $\mathbf{s} = \operatorname{Tokenize}(S)$.
    The autoregressive language model $f_l(\cdot)$ subsequently generates a sequence of output tokens $\mathbf{y} = \{y_1, \ldots, y_T\}$, conditioned on the encoded inputs and previously generated tokens.
    
    During decoding, the server measures the token-level uncertainty using an entropy function $g(\cdot)$. The average entropy across the output sequence is computed as:
    \begin{equation} 
        G = \frac{1}{T} \sum_{t=1}^T g(\mathbf{x}, \mathbf{s}, \mathbf{y}_{<t}). \label{eq:avg_entropy}
    \end{equation}
    If $G\ge \eta$, where $\eta$ denotes a predefined retransmission threshold, the server triggers a second stage and computes the relative attention map $A_{\text{rel}}$:
    \begin{equation}
        A_{\text{rel}} = \operatorname{ComputeRelativeAttention}(\mathbf{x}, \mathbf{s}).
    \end{equation}
    This map identifies the RoI that receives the highest model attention during generation.
    The server then determines a bounding box corresponding to the RoI and transmits its coordinate vector $\mathbf{b} \in \mathbb N^4$ to the edge device:
    \begin{equation}
        \mathbf{b} = \operatorname{FindBoundingBox}(A_{\text{rel}}).
    \end{equation}
    
    After receiving a retransmission request, the edge device crops the original image according to $\mathbf{b}$ to obtain a local image and transmits it to the server.
    The server encodes the local image as $\tilde{\mathbf{x}} = f_v(\tilde{I})$ and performs a second-stage inference using both the global and local features:
    \begin{equation}
        y'_t = f_l(\mathbf{x}, \tilde{\mathbf{x}}, \mathbf{s}, \mathbf{y}'_{<t}).
    \end{equation}
    During output generation in the second stage, the server evaluates the second entropy $G'$ as follows:
    \begin{equation} 
        G' = \frac{1}{T'} \sum_{t=1}^{T'} g(\mathbf{x}, \tilde{\mathbf{x}}, \mathbf{s}, \mathbf{y}'_{<t}),
    \end{equation}
    where $T'$ denotes the number of decoding steps of the second inference.
    By comparing the min-entropy values $G $ and $G'$, the server adaptively selects the most reliable result and transmits it to the edge device.     

   \section{Entropy-Aware Retransmission (EAR)} \label{sec:entropy}

    We incorporate inference uncertainty into the framework to adaptively balance communication cost and inference accuracy. 
    This framework accounts for the varying difficulty across different image-question pairs.
    For simple visual scenes or straightforward questions, inference based solely on the global image is often sufficient, eliminating the need for additional edge-server interaction.
    In contrast, complex scenes or semantically demanding questions require finer-grained visual information.
    This necessitates transmitting additional image data from the edge to the server, thereby increasing communication cost.

    A critical component of this framework is determining whether the initial inference is reliable or whether additional image information should be requested.
    Although the server cannot directly verify correctness, it can estimate the confidence of its prediction from the output probabilities of the LLM decoder.
    These probabilities are obtained by applying a softmax function to the logits of each generated token.
    Formally, the softmax output for the $t$-th token corresponds to the model’s predictive distribution: $p_\theta(y_t | \mathbf{x}, \mathbf{s}, \mathbf{y}_{<t}),$ where $\theta$ denotes the LLM parameters.
    We define an entropy function $g(\cdot)$ that maps the softmax distribution of the output token to a scalar entropy value.

    Because of the autoregressive nature of the LLM, the total uncertainty of a generated sequence can be estimated by aggregating the \emph{token-level} entropy values.
    The average predictive uncertainty is given by~\eqref{eq:avg_entropy}.
    If $G$ exceeds a retransmission threshold $\eta$, the server requests a retransmission to the edge device.
    This mechanism ensures that retransmissions occur only when necessary, maintaining reliability while conserving both communication and computational resources. 

    We measure inference uncertainty using three metrics: 1) Shannon entropy, 2) min-entropy, and 3) probability margin, each associated with its own threshold.
    The Shannon entropy, a widely used metric for quantifying uncertainty~\cite{Shannon1948mathematical}, is defined as
    \begin{align}
        g_s(\mathbf{x}, \mathbf{s}, \mathbf{y}_{<t}) = - \mathbb{E}_{p_\theta(y_t | \mathbf{x}, \mathbf{s}, \mathbf{y}_{<t})} \left[ \log_2 p_\theta(y_t | \mathbf{x}, \mathbf{s}, \mathbf{y}_{<t}) \right].
    \end{align}
    A high Shannon entropy indicates that the input sequence $\{\mathbf{x}, \mathbf{s}, \mathbf{y}_{<t}\}$ is challenging for VLM inference, implying lower confidence in the generated response.
    Therefore, if $ G_{s} \geq \eta_s$, the server requests a detailed local image to refine the inference result.
    
    The min-entropy provides a conservative measure of uncertainty~\cite{Konig2009operational,Kim2021on} and is defined as 
    \begin{equation}
        g_m(\mathbf{x}, \mathbf{s}, \mathbf{y}_{<t}) = -\log_2 \max_{y_t\in \mathcal Y}p_\theta(y_t | \mathbf{x}, \mathbf{s} , \mathbf{y}_{<t}),
    \end{equation}
    where $\mathcal{Y}$ denotes the set of all possible output tokens. 
    As in the Shannon entropy, if $G_{m} \geq \eta_m$, the server requests a local image for refinement.

    The final metric, probability margin, measures the gap between the probabilities of the most likely token $y_{t,1}$ and the second most likely token $y_{t,2}$~\cite{Settles2012active}.
    The probability margin is defined as 
   \begin{equation}
        g_p(\mathbf{x}, \mathbf{s}, \mathbf{y}_{<t}) = p_\theta(y_{t,1} | \mathbf{x}, \mathbf{s} , \mathbf{y}_{<t})-p_\theta(y_{t,2} | \mathbf{x}, \mathbf{s} , \mathbf{y}_{<t}).
    \end{equation}
    In contrast to the other uncertainty metrics, a smaller probability margin indicates lower confidence in the inference result.
    Accordingly, if $G_{p} \leq \eta_p$, the server requests a local image for refined inference.
    
    We evaluate the effectiveness of different uncertainty metrics by measuring the divergence between the entropy distributions of correct and incorrect samples.
    As shown in the experimental results in Section~\ref{sec:uncertainty_metric}, the min-entropy provides the most reliable criterion for requesting retransmission.
    Consequently, the entropy-aware image retransmission based on the min-entropy achieves higher communication efficiency at a given level of inference accuracy compared with alternative metrics.

    \section{Attention-Guided Collaborative Visual Cropping}\label{sec:collaborative_vicrop} 
    
\begin{figure}[t!] 
        \centering
        \includegraphics[width=0.42\textwidth]{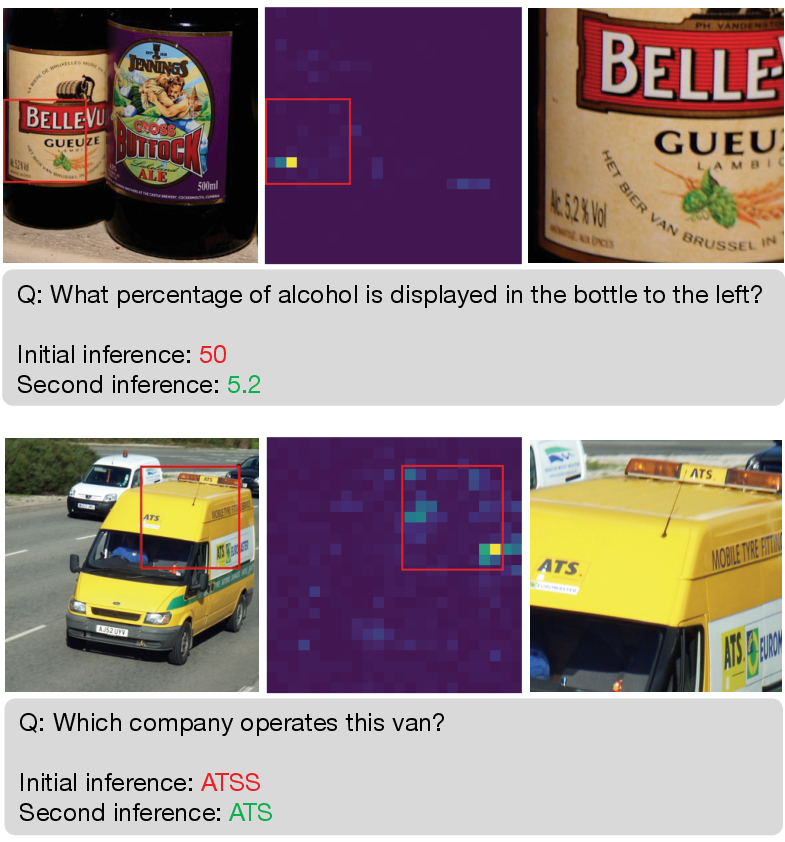}
        \caption{Visualization of the collaborative visual cropping. The left image shows the original input residing on the edge device; the middle image shows the server-computed relative attention map with the bounding box corresponding to the given question; and the right image shows the detail-preserved local image extracted from the edge device and transmitted to the server.} 
        \label{fig:collaborative_vicrop}
        \vspace{-4mm}
    \end{figure}

    The proposed framework incorporates ViCrop~\cite{Zhang2025mllms} into the collaborative edge–server inference for VLMs.
    By identifying semantically important regions in the original image only when necessary, the proposed method maintains the inference accuracy achieved by ViCrop while substantially reducing both communication cost and server-side computational overhead.
    Fig.~\ref{fig:collaborative_vicrop} shows visual examples of original edge-side images, server-generated relative attention maps, and resulting local RoI regions obtained for corresponding questions.   
    
    The proposed framework integrates ViCrop into its retransmission mechanism, executing it conditionally based on inference uncertainty. 
    Specifically, cropping is performed only when the min-entropy of the initial inference exceeds a retransmission threshold, ensuring that additional transmissions occur only when necessary.  
    Once activated, the server computes a relative attention map $A_{\text{rel}}$ and determines a bounding box $\mathbf{b}$ that localizes the RoI.  
    The server then sends a retransmission request along with $\mathbf{b}$ to the edge device.  
    Upon receiving the request, the edge device crops the original image accordingly and resizes the cropped region to match the input resolution of the vision encoder:
    \begin{equation}
        \tilde{I} = \operatorname{Resize}(\operatorname{Crop}(I, \mathbf{b})).
    \end{equation}
    The resulting local image $\tilde{I}$ is transmitted to the server, which performs a second inference using both the global and local images.  

    This collaborative extension of ViCrop transforms a purely accuracy-oriented cropping method in~\cite{Zhang2025mllms} into an integral component of the proposed framework.     
    By integrating entropy-aware retransmission with \emph{server-side RoI estimation} and \emph{edge-side cropping}, the proposed framework transmits semantically critical regions only when necessary.
    Experimental results presented in the following section demonstrate that this approach substantially reduces both communication and computation costs while maintaining inference accuracy.

    \section{Entropy-Guided Selective Refinement (SR)}\label{sec:selective_refinement} 
    The proposed framework selectively refines the final output by comparing the entropy evaluated in two inference stages to ensure the delivery of the most reliable result. 
    A secondary inference stage, triggered only if the initial entropy exceeds a retransmission threshold, leverages fine-grained visual information within the RoI.
    While these secondary inferences frequently yield higher accuracy, they do not consistently outperform initial inferences~\cite{Zhang2025mllms}.
    In certain instances, additional details provided during the second stage may act as distractions, potentially degrading the model's performance when the initial inference is correct.
    To mitigate this issue, we propose a selective refinement (SR) mechanism that compares the output uncertainties of the two stages.
    
\begin{figure}[t!] 
        \centering
        \includegraphics[width=0.4\textwidth]{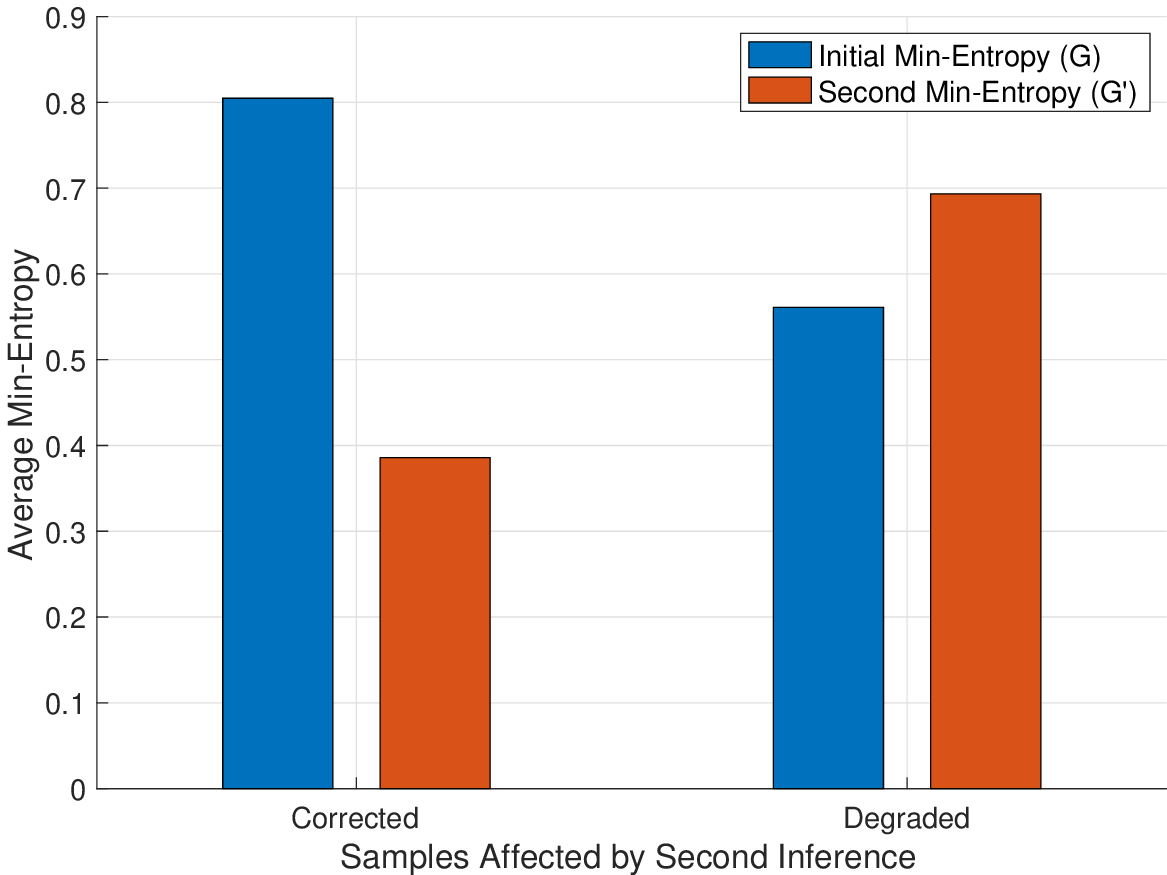}
        \caption{Comparison of average min-entropy between initial and second inferences for corrected and degraded sample groups. The contrast in entropy dynamics justifies the use of entropy difference ($\Delta G$) as a criterion for selective refinement.} 
        \label{fig:entropy_difference}
        \vspace{-4mm}
    \end{figure}

\begin{figure}[t!] 
        \centering
        \includegraphics[width=0.4\textwidth]{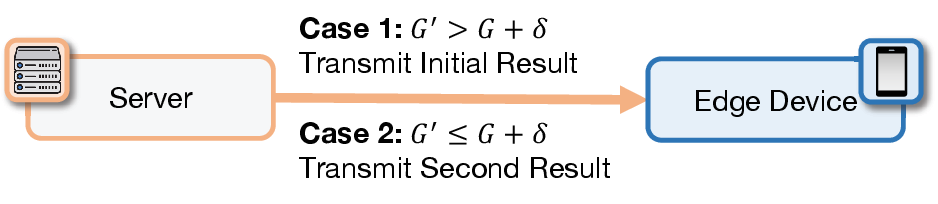}
        \caption{Illustration of the entropy-guided selective refinement strategy.} 
        \label{fig:selective_refinement}
        \vspace{-4mm}
    \end{figure}

    The proposed method is based on the observation that a higher predictive entropy in the second stage, relative to the initial entropy, is positively correlated with incorrect outputs.
    Fig.~\ref{fig:entropy_difference} shows that for successfully corrected samples, the entropy typically decreases. In contrast, samples that were initially correct but became incorrect after the second inference exhibit an increase in entropy.
    Motivated by these distinct tendencies, we develop a selective refinement mechanism illustrated in Fig.~\ref{fig:selective_refinement}.
    If $G^\prime$ exceeds $G+\delta$, the server regards the second inference as inaccurate and transmits the initial result to the client.
    Here, the refinement threshold $\delta$ acts as a tolerance level that reflects an inherent bias toward the higher reliability typically associated with the second inference.
    Otherwise, the second result is transmitted when the entropy difference $\Delta G=G'-
    G$ remains within this threshold.
    To establish a statistically sound decision boundary, the optimal threshold $\delta^*$ is determined by solving the following empirical risk minimization (ERM) problem:
    \begin{equation}
    \delta^* = \arg \min_{\delta} \left( \sum_{i:\Delta G_i \le \delta} \mathbb{I} (\mathbf{y}'_{i} \neq \mathbf{y}_{i}^{\text{GT}}) + \sum_{i:\Delta G_i > \delta} \mathbb{I} (\mathbf{y}_{i}\neq \mathbf{y}_{i}^{\text{GT}}) \right),\label{eq:erm}
    \end{equation}
    where $\mathbb{I}(\cdot)$ denotes the indicator function, $\mathbf{y}_i$, $\mathbf{y}'_i$, and $\mathbf{y}_i^{\text{GT}}$ represent the first and second inference result, and the ground-truth label of the $i$-th sample, respectively.
    Experimental evaluations consistently demonstrate that this entropy-guided selective refinement strategy effectively improves the trade-off between task accuracy and communication overhead.
    
    \section{Experimental Results}\label{sec:results}
        
    \subsection{Experiment Settings}

    We evaluate the proposed edge-to-server inference framework using two different types of state-of-the-art open-source VLMs: LLaVA-1.5 (Vicuna-7B)~\cite{Liu2024improved} and Qwen2.5-VL-3B~\cite{Bai2025qwen2}.
    These models primarily differ in the input resolution of their vision encoder. 
    Specifically, LLaVA-1.5 employs a vision encoder with a fixed input resolution of $336 \times 336$ pixels~\cite{Liu2024improved}.
    In contrast, Qwen2.5-VL employs a variable-resolution vision encoder. 
    To ensure a consistent comparison, we configure the vision encoder to generate an average of $144$ visual tokens, representing a pixel-area budget equivalent to a $336\times336$ pixel resolution~\cite{Bai2025qwen2}.
    Experiments are primarily conducted on the TextVQA benchmark~\cite{Singh2019towards}; to demonstrate the generalizability of our approach, we further extend our evaluation to the GQA~\cite{Hudson2019gqa} and VQAv2~\cite{Goyal2017making} validation datasets.
    
    In our experiments, we evaluate the tradeoff between the additional communication cost and task accuracy.
    The additional communication cost, $C_{\text{add}}$, is defined as follows:
    \begin{equation}
        C_{\text{add}}=\frac{\sum_i (B_{i,\text{total}}-B_{i,\text{global}})}{\sum_i B_{i,\text{global}}},
    \end{equation}
    where $B_{i,\text{global}}$ represents the baseline data volume required to transmit the $i$-th sample as a global image, and $B_{i,\text{total}}$ denotes the total amount of data transmitted for the $i$-th sample.
    Specifically, if the server requests local images for all data samples, i.e., the retransmission threshold $\eta$ is set to $0$, the additional communication cost equals $1$.
    On the other hand, when the server conducts inference only using global images, i.e., $\eta$ is set to $\infty$, the additional communication cost is $0$.
    For selective refinement (SR), the refinement threshold $\delta$ is set to 0.1 across the entire dataset and the benchmarks.

    \begin{figure}[t!] 
        \centering
        \subfloat[]{\includegraphics[width=0.4\textwidth]{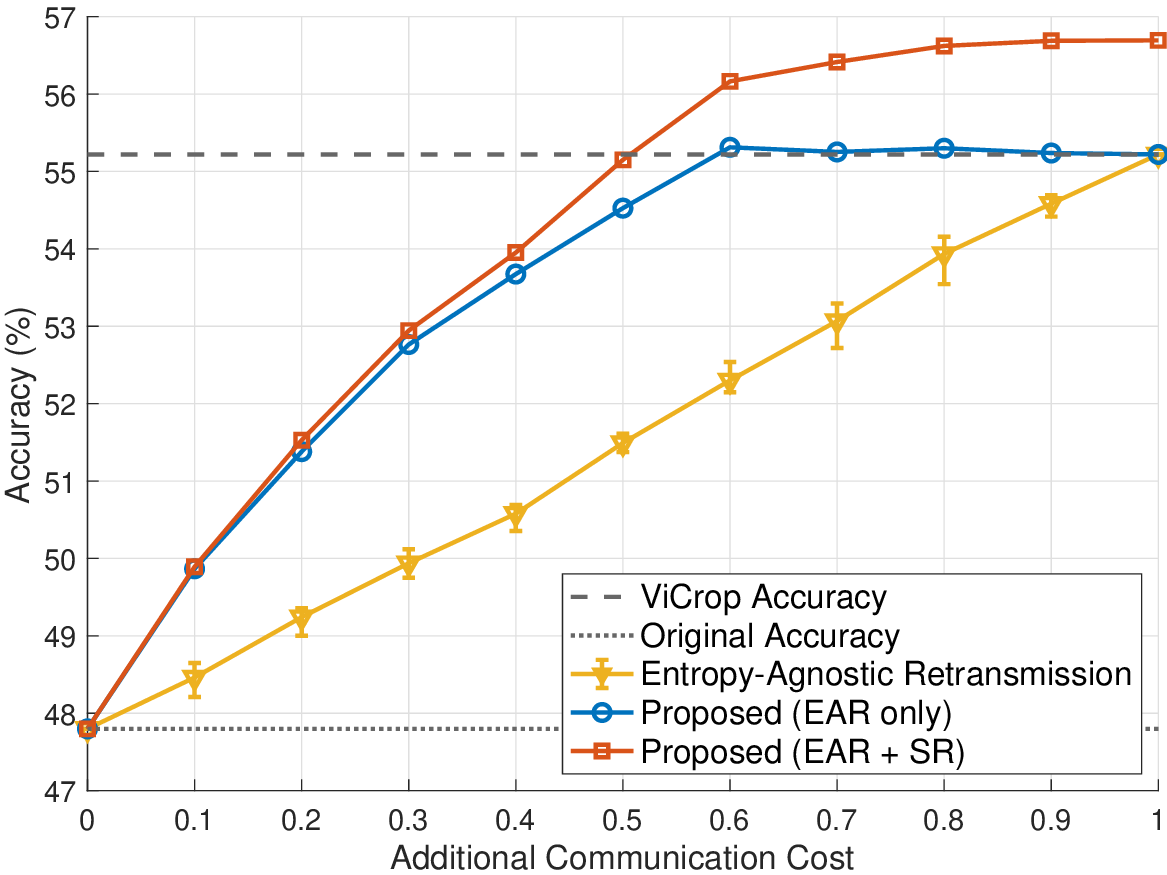}\label{fig:llava_textvqa}}
        \hfil
        \subfloat[]{\includegraphics[width=0.4\textwidth]{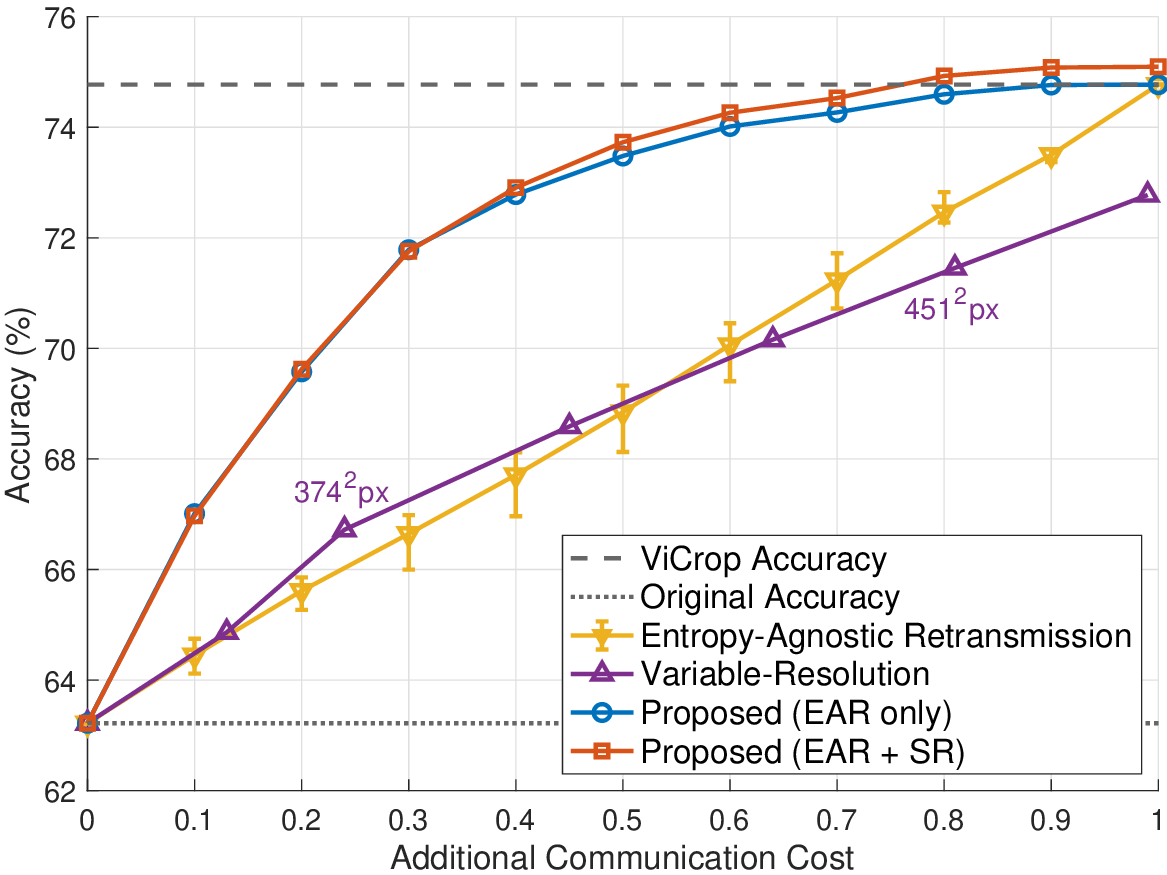}\label{fig:qwen_textvqa}}
        \caption{Tradeoff between communication cost and inference accuracy for (a) LLaVA-1.5-7B and (b) Qwen2.5-VL-3B on TextVQA benchmark. The uncertainty metric used is min-entropy. In (b), the $i^2$px notation along the variable-resolution curve indicates that the vision encoder operates at an average input resolution of $i \times i$ pixels. The proposed framework achieves a more favorable communication--accuracy tradeoff compared to baselines.}
        \label{fig:communication_efficiency}
        \vspace{-4mm}
    \end{figure} 
    
    \subsection{Communication Efficiency}\label{sec:main_result}

    We first evaluate the effectiveness of the proposed framework in terms of communication costs.
    Fig.~\ref{fig:communication_efficiency} shows the tradeoff between communication cost and inference accuracy obtained by LLaVA-1.5 and Qwen2.5-VL on TextVQA benchmarks.\footnote{The LLaVA-1.5 accuracy on TextVQA is lower than reported in \cite{Liu2024improved} because we follow the evaluation protocol of \cite{Zhang2025mllms} without utilizing additional OCR tokens.}
    The dotted line labeled \emph{Original accuracy} denotes the accuracy when only single-stage inference is performed, whereas \emph{ViCrop accuracy} line represents the accuracy when all samples are retransmitted in the two-stage scheme.
    We also include \emph{Entropy-agnostic retransmission} as a baseline, in which local images identified by ViCrop are retransmitted without accounting for sample-level entropy.
    Furthermore, in Fig.~\ref{fig:communication_efficiency}(b), \emph{Variable-resolution} refers to a single-stage inference baseline configured with a variable, higher-resolution setting.
    
    The results demonstrate that the proposed framework achieves higher accuracy at lower communication cost.
    As shown in Fig.~\ref{fig:communication_efficiency}(a) and Fig.~\ref{fig:communication_efficiency}(b), the proposed method matches the ViCrop accuracy while achieving a reduction in additional communication costs by \SI{50}{\%} and \SI{24}{\%}, respectively.
    Furthermore, our approach indicates enhanced performance across all communication costs compared to the entropy-agnostic retransmission baseline.
    Notably, as shown in Fig.~\ref{fig:communication_efficiency}(b), the proposed method offers a more favorable accuracy-cost tradeoff than the variable-resolution baseline.
    These results imply that our mechanism effectively allocates the communication budget by prioritizing task-relevant and semantically significant information, thereby enhancing the efficiency of variable-resolution VLMs.

    \begin{figure}[t!] 
        \centering
        \includegraphics[width=0.4\textwidth]{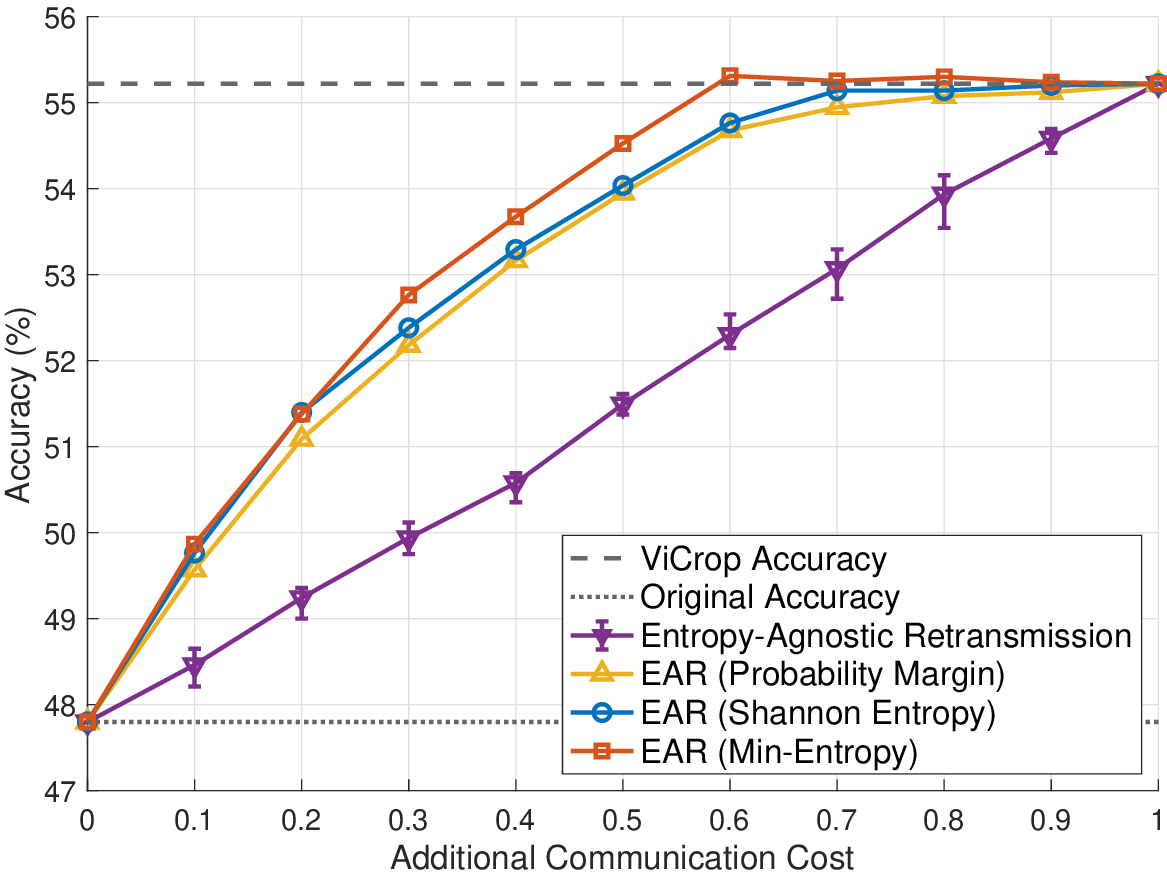}
        \caption{Comparison of uncertainty metrics on TextVQA benchmark using LLaVA-1.5-7B. The min-entropy-based retransmission strategy shows the best accuracy-communication cost tradeoff compared with alternative metrics.}
        \label{fig:entropy_metric}
        \vspace{-4mm}
    \end{figure}

  \subsection{Choice of Uncertainty Metric}\label{sec:uncertainty_metric}
    
    Next, we investigate which uncertainty metric provides the most effective tradeoff between task accuracy and communication cost.
    As shown in Fig.~\ref{fig:entropy_metric}, we compare the performance of several uncertainty-based retransmission strategies--Shannon entropy, min-entropy, and probability margin--against an entropy-agnostic retransmission baseline.
    The results show that EAR substantially outperforms entropy-agnostic retransmission, achieving accuracy comparable to the ViCrop accuracy with only \SI{60}{\%} of the additional communication cost. 
    Among the uncertainty metrics, min-entropy consistently provides the best performance across most operating points.

    \begin{table}[t!]
        \centering
        \renewcommand{\arraystretch}{1.2}
        \caption{Statistical Distances between Entropy Distributions of Correct and Incorrect Samples}
        \label{tab:distance}
        \begin{tabular}{lccc}
        \hline
        \textbf{} & \textbf{Min-Entropy} & \textbf{Shannon Entropy} & \textbf{Prob. Margin} \\
        \hline
        Overlap ($\downarrow $)  & $\mathbf{0.47}$ & $0.54$ & $0.49$ \\
        $D_B$ ($\uparrow$)    & $\mathbf{0.33}$ & $0.27$ & $0.24$ \\
        \hline
        \end{tabular}
        \vspace{-4mm}

    \end{table}

    Table~\ref{tab:distance} reports the distance between the entropy distributions of correct and incorrect predictions, computed using three different uncertainty estimation methods.
    To quantify the separability between distributions, we employ two metrics: the distribution overlap and the Bhattacharyya distance ($D_B$)~\cite{Bhattacharyya1946on}. 
    These metrics, which measure the distance between two PMFs $p(x)$ and $q(x)$ are defined as:
    \begin{align}
        \operatorname{Overlap}(p, q)& = \sum_x \min(p(x),q(x)),\\
        D_B(p, q) &= -\ln\sum_x \sqrt{p(x)q(x)}.
    \end{align}
    
    A consistent ordering is observed between Bhattacharyya distances and the accuracy gains of each uncertainty measure.
    Specifically, the probability margin yields the smallest distance and, correspondingly, the smallest accuracy improvement.
    In contrast, min-entropy, which shows the largest distance, achieves the best performance.
    Taken together with the results in Fig.~\ref{fig:entropy_metric}, these findings confirm that the min-entropy is the most discriminative uncertainty measure for our collaborative inference framework.

    \subsection{Uncertainty Aggregation Strategy}\label{sec:aggregation}
    
    \begin{figure}[t!] 
        \centering
        \subfloat[]{\includegraphics[width=0.4\textwidth]{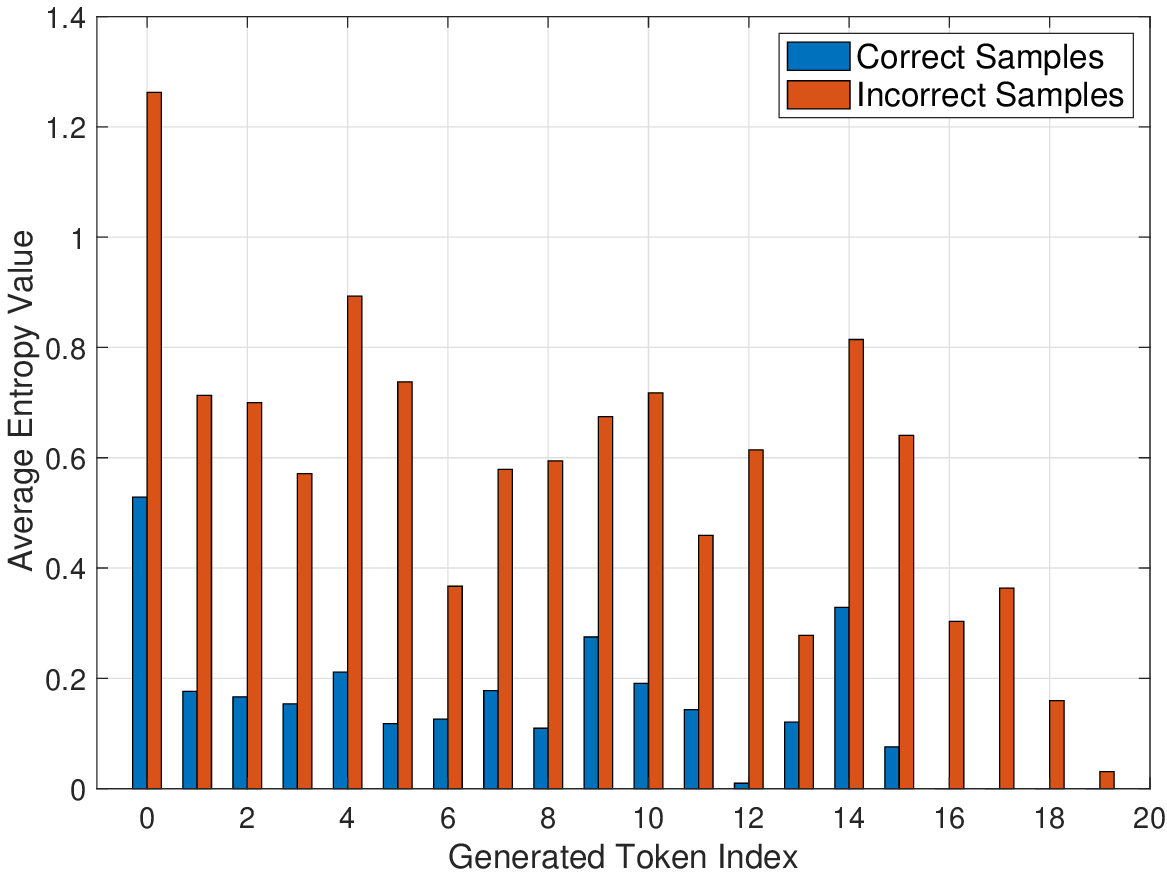}\label{fig:token_entropy}}
        \hfil
        \subfloat[]{\includegraphics[width=0.4\textwidth]{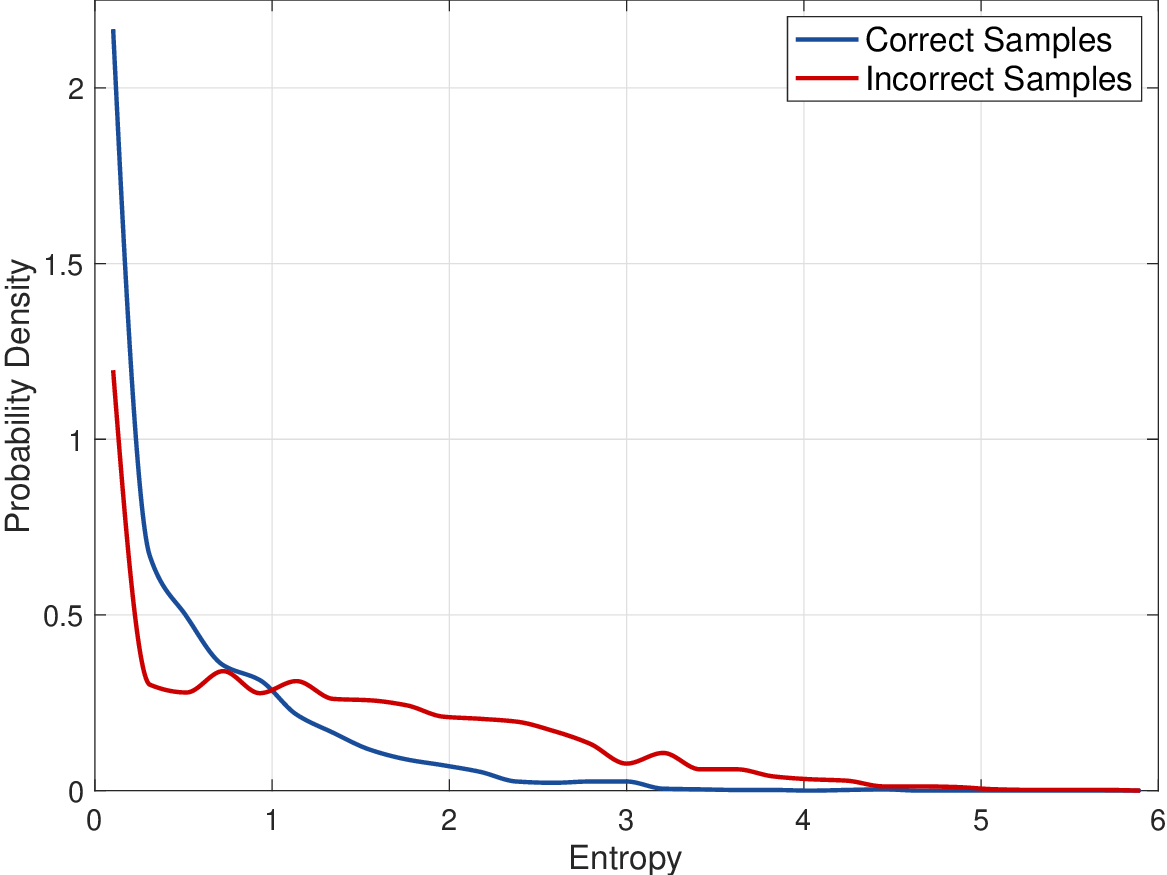}\label{fig:starting_distribution}}
        \hfil
        \subfloat[]{\includegraphics[width=0.4\textwidth]
        {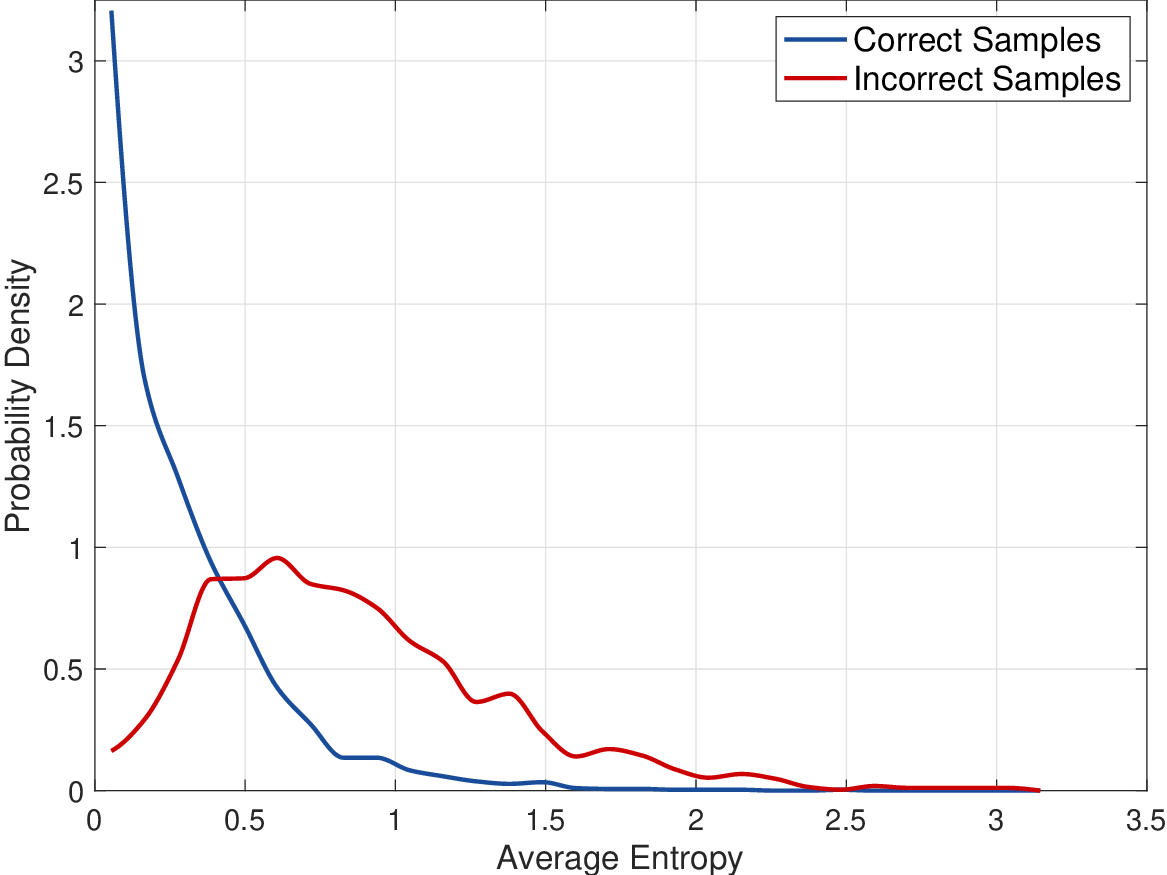}\label{fig:full_entropy_distribution}}
        \caption{Analysis of min-entropy aggregation on TextVQA benchmark. (a) Average min-entropy at each generation step for correct and incorrect samples, (b) min-entropy distribution of the starting token, and (c) min-entropy distribution of the full-sequence average.}
        \label{fig:aggregation_analysis}
        \vspace{-4mm}
    \end{figure} 
    
    We analyze the effectiveness of token-level entropy as a measure of VLM inference uncertainty and investigate how to aggregate token-wise values into a sequence-level uncertainty score. 
    First, we statistically compare the token-level min-entropy of correct and incorrect predictions produced by the LLaVA-1.5-7B model on the TextVQA benchmark.
    Fig.~\ref{fig:aggregation_analysis}(a) shows the average min-entropy value at each generation step, where the maximum output length is set to $20$ tokens. 
    Incorrect samples consistently exhibit higher min-entropy across all token positions. 
    In contrast, correct samples maintain lower min-entropy values, indicating more confident predictions throughout the sequence.
    This clear separation shows that token-level min-entropy provides meaningful cues for distinguishing correct from incorrect predictions.

    \begin{table}[t!]
        \centering
        \renewcommand{\arraystretch}{1.2}
        \caption{Statistical Distances between Entropy Distributions of Correct and Incorrect Samples under Token Aggregation}
        \label{tab:distance_token_num}
        \begin{tabular}{lcc}
        \hline
        \textbf{}  & \textbf{Starting Token} & \textbf{Full Token Average} \\
        \hline
        Overlap ($\downarrow $)  & $0.63$ & $\mathbf{0.47}$ \\
        $D_B$ ($\uparrow$) & $0.19$ & $\mathbf{0.33}$ \\
        \hline
        \end{tabular}
    \end{table}

    \begin{figure}[t!] 
        \centering
        \subfloat[]{\includegraphics[width=0.4\textwidth]{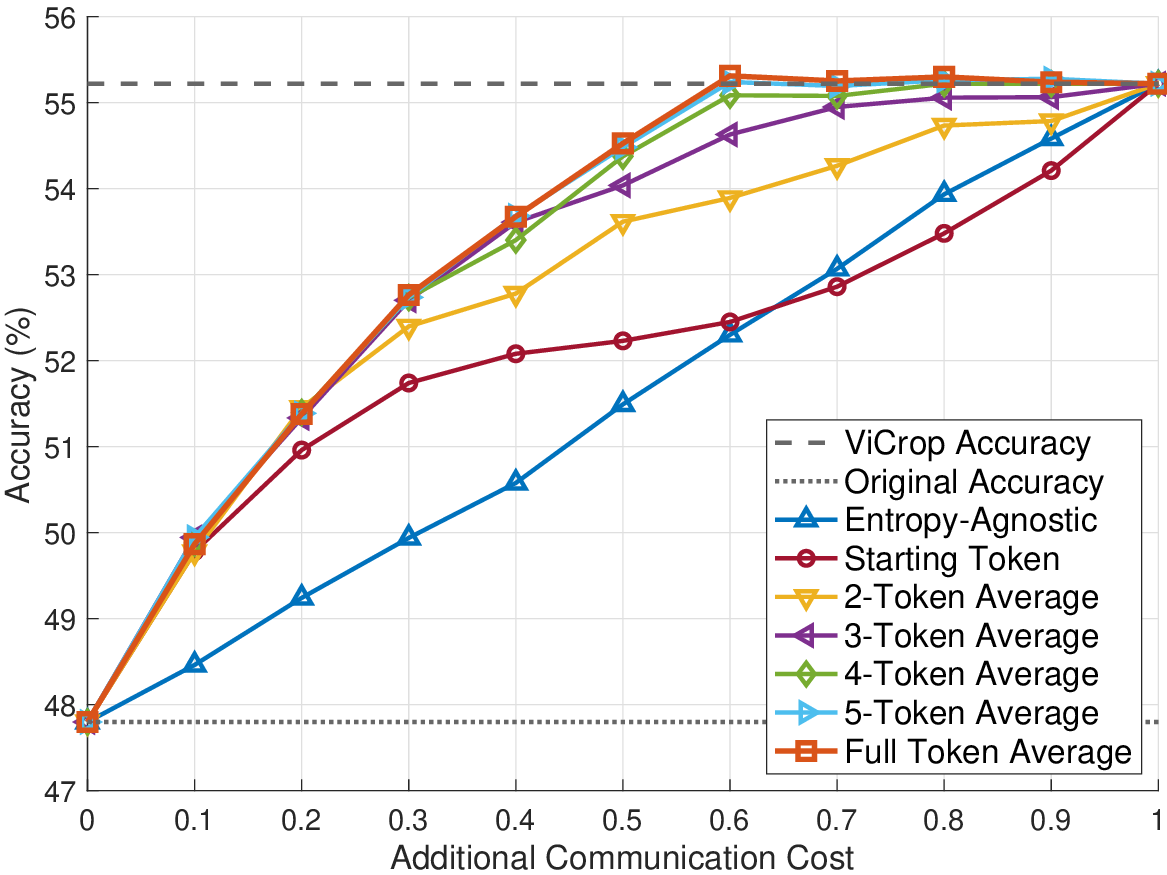}\label{fig:token_aggregation}}
        \hfil
        \subfloat[]{\includegraphics[width=0.4\textwidth]{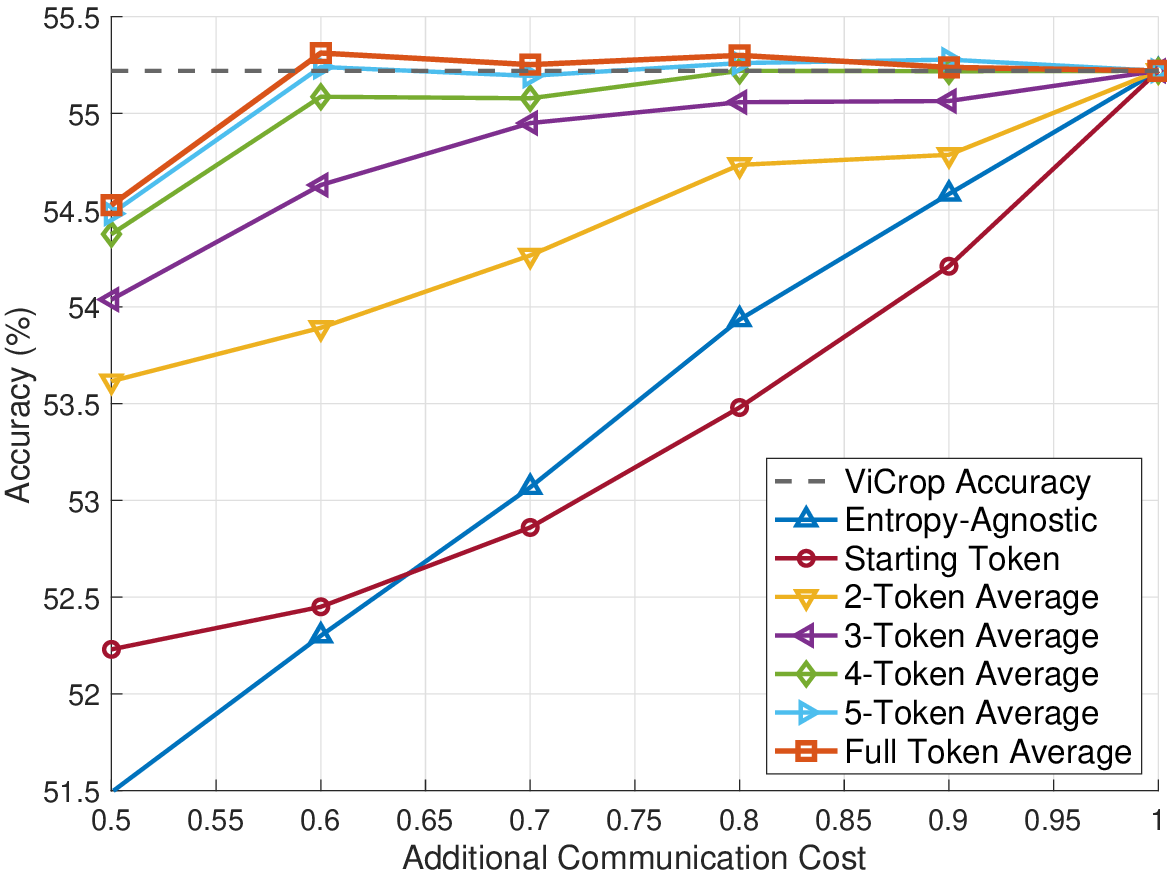}\label{fig:token_aggregation_zoom}}
        \caption{Comparison of tradeoff curves for different token aggregation lengths. (a) shows the overall tradeoff, while (b) zooms in on the region of interest, focusing on the high-accuracy regime.}
        \label{fig:uncertainty_method}
        \vspace{-4mm}
    \end{figure}

    Fig.~\ref{fig:aggregation_analysis}(b) and Fig.~\ref{fig:aggregation_analysis}(c) compare the min-entropy distributions of the starting token and the full-sequence average, respectively.
    In Fig.~\ref{fig:aggregation_analysis}(b), a considerable fraction of incorrect samples show the min-entropy values close to zero, which is less evident in Fig.~\ref{fig:aggregation_analysis}(c).
    Table~\ref{tab:distance_token_num} quantifies this effect: starting-token min-entropy exhibits higher overlap and lower Bhattacharyya distance, reflecting weaker discriminative power compared to full-sequence averaging.
    These results confirm that aggregating the min-entropy values across the entire sequence provides a more reliable uncertainty estimate, improving the separation between correct and incorrect predictions.
    Therefore, full-sequence min-entropy aggregation--rather than relying solely on a starting token alone--provides a more accurate uncertainty estimate for the proposed collaborative inference.

    To further examine the influence of the aggregation length, we analyze how task accuracy varies with the number of tokens included in the entropy average.
    Fig.~\ref{fig:uncertainty_method} shows the tradeoff between task accuracy and communication cost as a function of aggregation length, where the ``$i$-token average'' refers to using the min-entropy averaged over the first $i$ generated tokens to determine whether retransmission is necessary.
    Notably, averaging over only the first five tokens achieves performance comparable to full-sequence aggregation, indicating that a limited number of initial tokens often suffices to characterize inference uncertainty without requiring full-sequence generation in the first stage.

   \subsection{Effectiveness Across Diverse Benchmarks }\label{sec:application}

    \begin{figure}[t!] 
        \centering
        \subfloat[]{\includegraphics[width=0.23\textwidth]{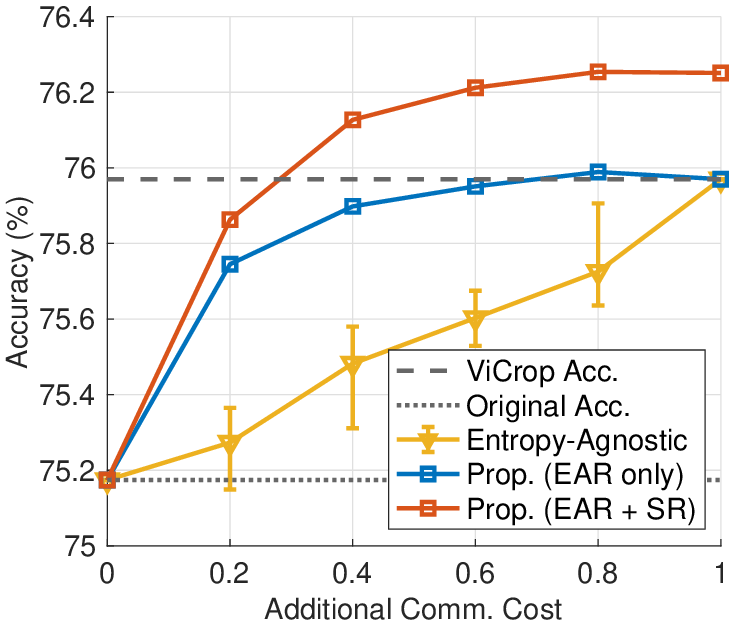}\label{fig:llava_vqav2}}
        \hfil
        \subfloat[]{\includegraphics[width=0.23\textwidth]{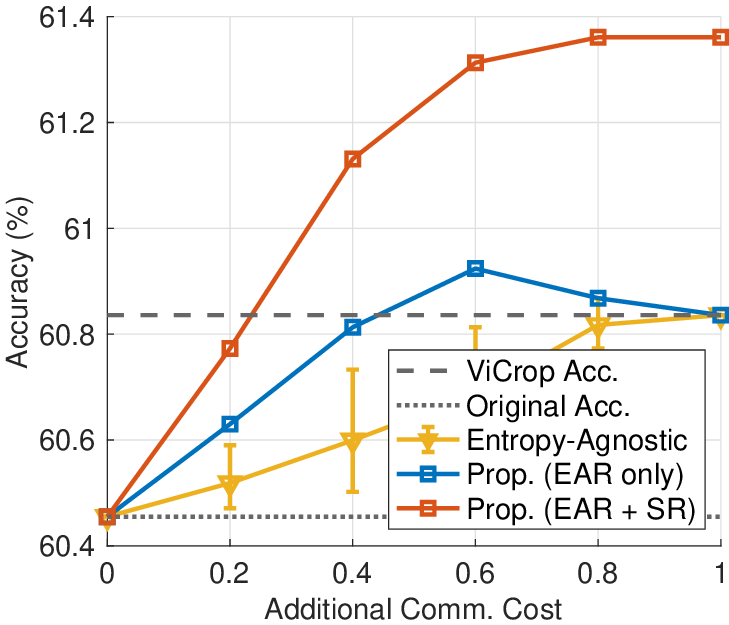}\label{fig:llava_gqa}}
        \hfil
        \caption{Tradeoff between additional communication cost and task accuracy on LLaVA-1.5-7B across different benchmarks: (a) VQAv2 and (b) GQA.
        }
        \label{fig:different_benchmark_llava}
        \vspace{-4mm}
    \end{figure}

    \begin{figure}[t!] 
        \centering
        \subfloat[]{\includegraphics[width=0.23\textwidth]{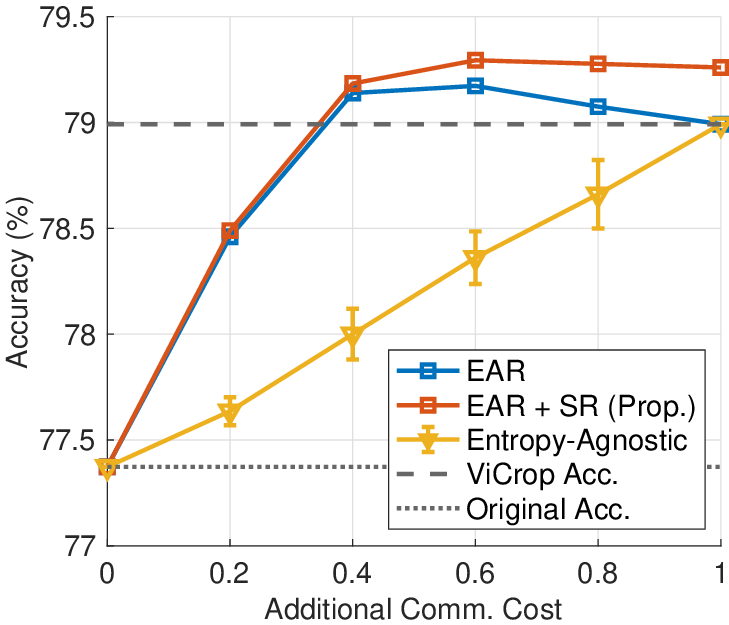}\label{fig:qwen_vqav2}}
        \hfil
        \subfloat[]{\includegraphics[width=0.23\textwidth]{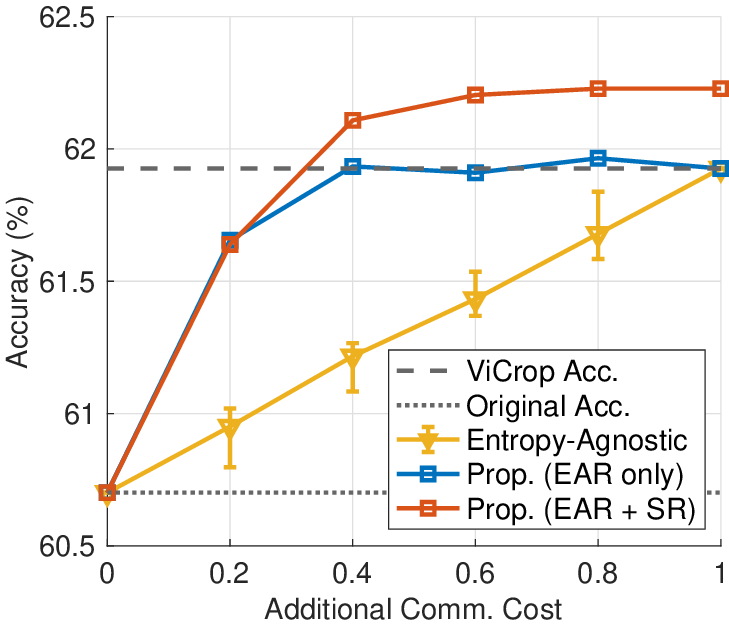}\label{fig:qwen_gqa}}
        \hfil
        \caption{Tradeoff between additional communication cost and task accuracy on Qwen2.5-VL-3B across different benchmarks: (a) VQAv2 and (b) GQA.}
        \label{fig:different_benchmark_qwen}
    \end{figure} 
    We demonstrate the practical usefulness of EAR and the SR mechanisms by applying them to additional VQA benchmarks.
    Fig.~\ref{fig:different_benchmark_llava} and Fig.~\ref{fig:different_benchmark_qwen} illustrate the generalizability of our proposed entropy-aware retransmission strategy across different VQA benchmarks: VQAv2 and GQA~\cite{Goyal2017making,Hudson2019gqa}.
    The proposed method attains equivalent accuracy to ViCrop while consuming approximately \SI{30}{\%} of the additional communication cost across the evaluated model and the benchmarks. Furthermore, our approach consistently outperforms entropy-agnostic baselines across the entire range of communication overheads, validating its effectiveness across diverse VQA tasks.

    \begin{figure}[t!] 
        \centering
        \includegraphics[width=0.4\textwidth]{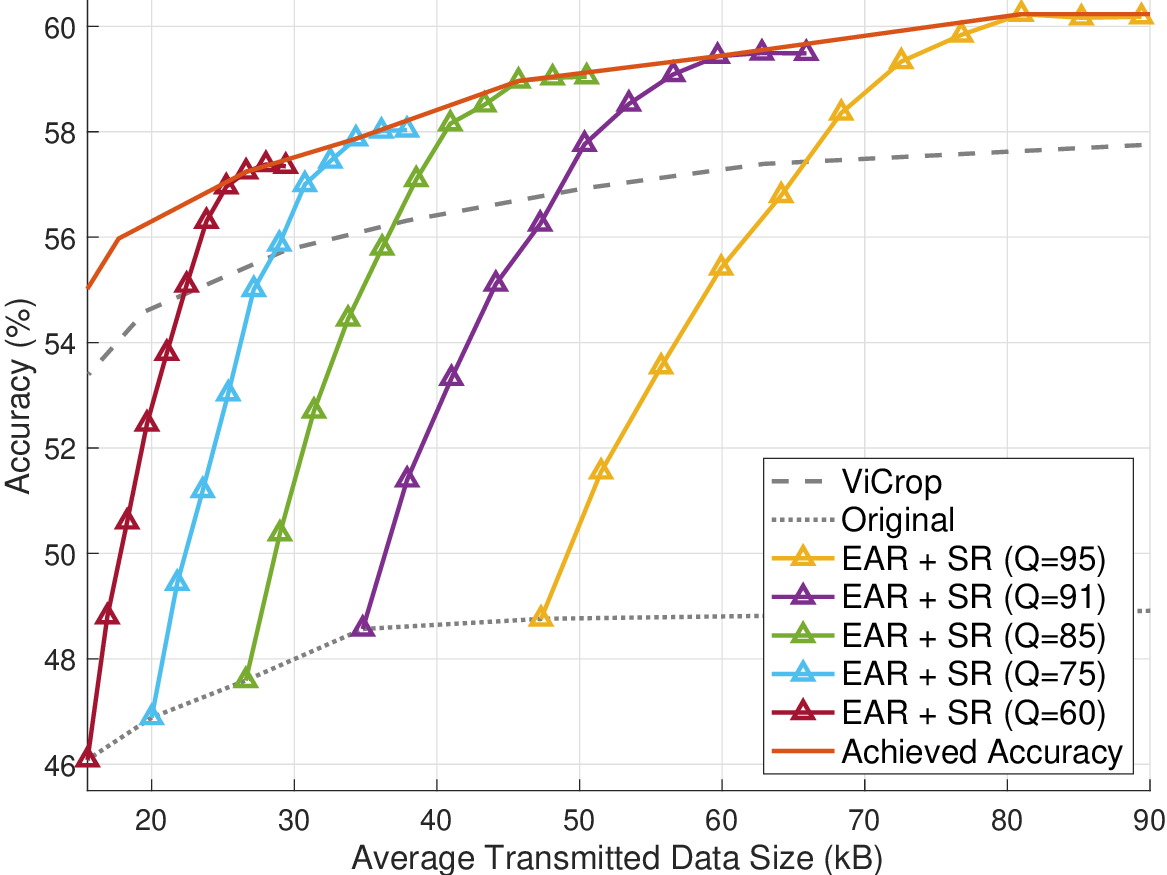}
        \caption{Additional gain achieved by combining the proposed method with JPEG image compression. Results are evaluated using LLaVA-1.5-7B on the TextVQA benchmark.}
        \label{fig:compression}
        \vspace{-4mm}
    \end{figure}
   \subsection{Evaluation on Selective Refinement Strategy}\label{sec:eval_sr}

    \begin{figure}[t!] 
        \centering
        \includegraphics[width=0.4\textwidth]{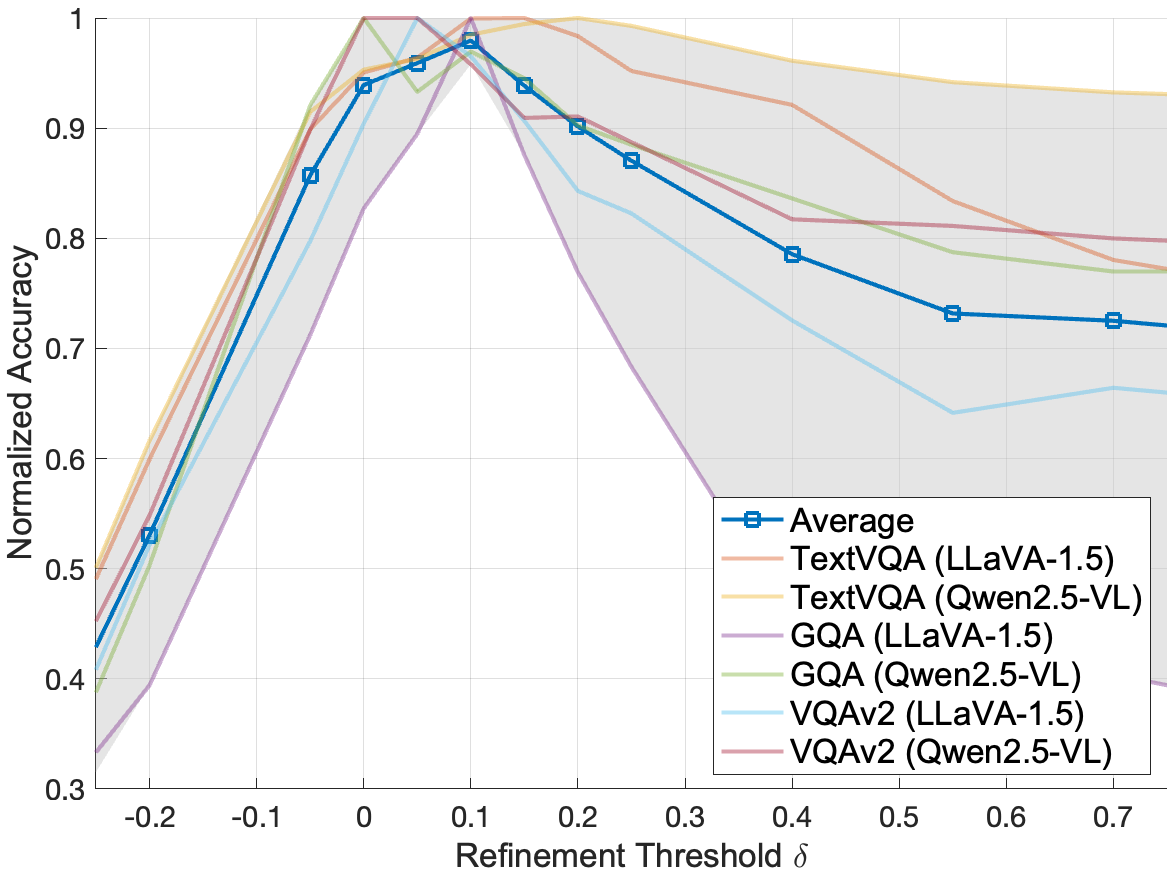}
        \caption{Comparison of normalized task accuracy as a function of $\delta$ across different benchmarks and VLM architectures.}
        \label{fig:eval_sr}
        \vspace{-4mm}
    \end{figure}
        
        In this section, we evaluate the impact of the selective refinement strategy on the overall system performance.
        As formulated in the ERM problem in~\eqref{eq:erm}, the optimal refinement threshold $\delta^*$ is determined at the crossing point where the number of corrected samples in the second inference matches that of the degraded samples, representing the statistical balance between performance gain and degradation.
        However, in practical deployments where ground-truth labels are unavailable, determining the absolute optimal threshold $\delta^*$ for each individual dataset is often infeasible.
        To address this, our empirical analysis demonstrates that a fixed threshold serves as a highly effective and practical alternative.
        As illustrated in Fig.~\ref{fig:eval_sr}, the average normalized accuracy reaches a prominent peak at $\delta=0.1$ across various datasets and VLM architectures, with performance tapering off as $\delta$ deviates from this value.
        
        Table~\ref{tab:refinement} compares the optimal threshold $\delta^*$ with a fixed threshold of $\delta=0.1$ across various datasets and models.
        The accuracy values reported herein are achieved at an additional communication cost of $1$.
        Notably, the resulting accuracy degradation from using a fixed $\delta=0.1$ is negligible compared to the substantial accuracy gains provided by our proposed method.
        Based on the findings in Fig.~\ref{fig:eval_sr} and Table~\ref{tab:refinement}, $\delta=0.1$ is identified as a robust and near-optimal threshold. This configuration enables significant accuracy gains without requiring per-dataset optimization.

        \begin{table}[t!]
            \centering
            \renewcommand{\arraystretch}{1.2}
            \setlength{\tabcolsep}{4pt}
            \caption{Comparison of Accuracy at the Optimal Threshold $\delta^*$ and Performance Degradation with a Universal Threshold $\delta=0.1$}
            \label{tab:refinement}
            \small 
            \begin{tabular}{llccc} 
            \hline
            \textbf{Model} & & \textbf{TextVQA} & \textbf{GQA} & \textbf{VQAv2} \\
            \hline
            \multirow{4}{*}{LLaVA-1.5} 
            & $\delta^*$  & 0.11 & 0.09 & 0.05 \\
            & Acc. on $\delta^*$  & 56.69 & 61.37 & 76.28 \\
            & Acc. on $\delta=0.1$ & 56.68 & 61.36 & 76.25 \\
            & Acc. Drop    & 0.01 & 0.01 & 0.03 \\
            \hline
            \multirow{4}{*}{Qwen2.5-VL} 
            & $\delta^*$  & 0.22 & 0.00 & 0.04 \\
            & Acc. on $\delta^*$  & 75.24 & 62.23 & 79.35 \\
            & Acc. on $\delta=0.1$ & 75.09 & 62.19 & 79.26 \\
            & Acc. Drop    & 0.15 & 0.04 & 0.09 \\
            \hline
            \end{tabular}
            \vspace{-4mm}
        \end{table}
 
    \subsection{Compatibility with Image Compression}  

    The proposed method also shows strong compatibility with image compression techniques, enabling additional reductions in communication cost.
    Fig.~\ref{fig:compression} shows the accuracy–communication tradeoff under different JPEG compression qualities.
    Specifically, \emph{EAR + SR (Q=$i$)} denotes the setting where both the global image and its corresponding local crop (if applicable) are compressed with a quality factor of $i$ prior to edge-to-server transmission in the two-stage inference process.
    The leftmost point of each colored curve represents the baseline scenario where only the compressed global image is transmitted.
    The subsequent points to the right illustrate the tradeoff between accuracy and communication cost, where the transmission ratio of the local image is governed by the retransmission threshold $\eta$.
    By integrating the proposed method with image compression, further communication savings can be obtained; for instance, the average transmitted data size is reduced from \SI{63}{kB} to \SI{28}{kB} while maintaining a task accuracy of \SI{57.4}{\%}, compared with the data size required by ViCrop.
    These results underscore the feasibility of the proposed method for seamless integration into existing communication frameworks, ensuring high compatibility with existing standardized protocols.
    
    \subsection{End-to-End Latency Analysis}\label{sec:latency} 
    \begin{figure}[t!] 
        \centering
        \includegraphics[width=0.4\textwidth]{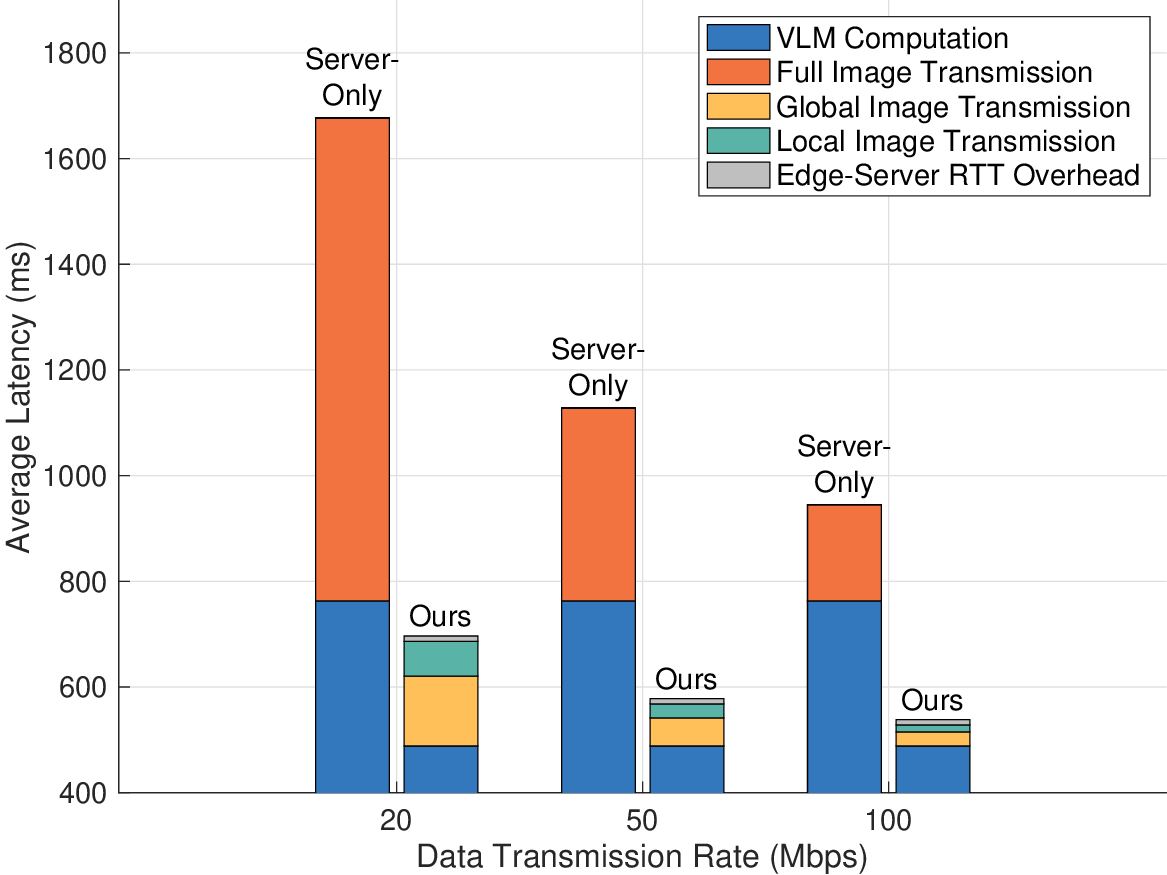}
        \caption{Comparison of average end-to-end latency between server-only inference and proposed method using LLaVA-1.5-7B and TextVQA across different data transmission rates. The VLM computation latency is evaluated on an NVIDIA A6000 GPU.}
        \label{fig:latency}
        \vspace{-4mm}
    \end{figure}
    
    Finally, we investigate the end-to-end inference latency to evaluate the practical efficiency of the proposed framework.    Fig.~\ref{fig:latency} compares the average end-to-end inference latency of the proposed method against a server-only baseline across different transmission rates ($20,50,$ and \SI{100}{Mbps}).
    The comparison is performed at a fixed accuracy level, which corresponds to the operating point where the additional communication cost is $0.5$ in Fig.~\ref{fig:communication_efficiency}(a).
    In the server-only inference, the edge device transmits the original image, and ViCrop is executed on the server.
    The proposed method offers significant advantages in terms of both VLM computation and edge-server communication latency.
    Specifically, the computation latency is reduced as the frequency of performing the secondary inference stage is effectively halved.
    Furthermore, by reducing the transmitted pixel data, our approach substantially decreases transmission latency.
    For instance, the raw original image size of \SI{2286}{kB} on average is condensed to a mere \SI{330}{kB} for both global and local images.
    Despite the additional edge-server round-trip time (RTT)--modeled after empirical 5G data~\cite{Ficzere2021real}--the proposed collaborative protocol achieves lower overall latency by significantly reducing both transmission data volume and VLM computational requirements.
    
    \section{Conclusion}\label{sec:conclusion}
    
    We proposed a novel collaborative edge-to-server inference framework that capitalizes on the capabilities of pre-trained VLMs.
    In the initial inference stage, the server performs inference on a global image transmitted from the edge device and identifies task-relevant regions based on the model's internal attention.
    The reliability of this initial inference is assessed using the min-entropy of output tokens. 
    When the min-entropy exceeds the retransmission threshold, a second-stage inference is triggered using both global and local images.
    By comparing the uncertainties obtained in two stages, we selectively refine the final answer, ensuring a high level of accuracy and communication efficiency.
    
    Experimental results show that the proposed framework substantially reduces communication and computation overhead while enhancing inference accuracy relative to baseline methods.
    The proposed method remains effective across different VQA tasks and VLM architectures.
    Furthermore, its combination with existing image compression techniques enables additional communication savings. 
    Future work will explore extending the framework to more complex communication settings, including robust transmission and multi-access environments.
    
    \appendices
	
    \bibliographystyle{IEEEtran}
    \bibliography{abrv,mybib}

\end{document}